# Hierarchical Physics-Embedded Learning for Partially Known Spatiotemporal Dynamics

Xizhe Wang[1,2,3*], Xiaobin Song[1,2,3*], Hongbo Zhao[4], Qingshan Jia[1,2,3], Qianchuan Zhao[1,2,3], Hao Sun[5], and Benben Jiang[1,2,3†]

*1 Department of Automation, BNRist, Tsinghua University, Beijing 100084, China*

*2 Beijing Key Laboratory of Embodied Intelligence Systems, Beijing 100084, China*

*3 Institute for Embodied Intelligence and Robotics, Tsinghua University, Beijing 100084, China*

*4 Department of Physics, Department of Chemistry and Biochemistry, University of California, San Diego, CA 92093, USA*

*5 Gaoling School of Artificial Intelligence, Renmin University of China, Beijing 100086, China*

** These authors contributed equally to this work*

*† Corresponding authors*

## Abstract

Partial physical knowledge—governing structures known, constitutive relations or their combinations not—pervades spatiotemporal systems. Existing scientific machine-learning paradigms learn evolution largely from data, impose equations as soft constraints, or hard-code physical terms into network updates; none exploits knowledge of this form. Here we introduce the hierarchical physics-embedded adaptive Fourier neural operator (HPE-AFNO), encoding such knowledge as computational architecture rather than penalizing or appending it: a first level learns or embeds fundamental physical expressions as intermediate representations, and a second level learns or embeds their governing combination, with adaptive Fourier layers capturing nonlocal, high-order couplings at each level. We prove a hierarchical error decomposition—embedding known components removes or shrinks their terms—and a parameter-complexity advantage: when the hierarchy aligns with the compositional structure of the dynamics, the number of learnable Fourier parameters sufficient for a prescribed accuracy grows strictly more slowly than for a single-level operator. Across canonical phase-field systems and experimental hydrofoil-wake data, HPE-AFNO reduces long-horizon extrapolation errors by 39–72% relative to state-of-the-art physics-encoded and neural-operator baselines, while preserving physically meaningful morphology, energetic consistency, and spectral structure, and maintaining robust performance under sparse and noisy observations. The separated intermediate representations further enable symbolic recovery of unknown constitutive relations in partially specified PDEs. These results establish hierarchical physics embedding as a theoretically grounded route to prediction and discovery when governing laws are neither fully known nor absent, but partially known and compositionally organized.

## 1 Introduction

Partial differential equations (PDEs) provide compact and interpretable descriptions of spatiotemporal dynamics across natural and engineered systems. They encode transport[1], reaction[2,3], phase separation[4,5], pattern formation[6,7], wave propagation[8,9] and fluid motion[10], and remain central to predictive modeling in physics, materials science, biology and engineering[11,12]. In many scientific problems, however, the governing equations are only partially known. A conservation law, a differential operator, an interfacial-energy term or a transport descriptor may be available from physical reasoning, whereas the constitutive relation, closure term, nonlinear reaction, mobility function or effective coupling that completes the dynamics remains unknown. The central difficulty in such problems is not the absence of physical knowledge but its incompleteness—and, critically, the fact that what is known and what is unknown are organized compositionally within the structure of the equation itself.

This form of incomplete physical knowledge is pervasive in far-from-equilibrium and pattern-forming systems. In phase-field dynamics[13,14], the variational or differential structure is often known, while the concentration-dependent mobility, free-energy derivatives or nonlinear reaction terms that determine the quantitative evolution are not. In measured and computational fluid flows[15,16], quantities such as vorticity and nonlinear convective descriptors provide meaningful information about rotation and transport, yet they do not by themselves close a temporal update for the velocity field, because pressure, boundary effects, viscous dissipation, three-dimensional disturbances and measurement noise remain entangled. Comparable situations arise in reaction–diffusion systems[17], electrochemical materials[18] and biological patterning[19]. What these systems share is the need for learning frameworks that exploit the physical structure that is available without forcing unresolved dynamics into an over-specified equation.

Existing scientific machine-learning paradigms address this regime only partially. Purely data-driven neural operators, including Fourier neural operators[20], adaptive Fourier neural operators[21] and DeepONet[22], can learn mappings between observed states and future dynamics with high expressive capacity. However, their physical structure is implicit, which can increase data requirements and lead to nonphysical drift during long-horizon extrapolation. Physics-informed neural networks impose known PDE residuals as soft constraints in the training loss[23-25], but the residual must be computable, and hence the governing equation must be essentially complete; when key terms are unknown, the constraint cannot be formed. Physics-encoded architectures such as PeRCNN[26] and PhyCRNet[27] hard-wire known differential operators into neural state-update rules, improving physical consistency when the available knowledge takes the form of predefined local terms. They do not, however, represent the compositional organization of scientific equations, in which elementary physical expressions and the rules that combine them may each be known, unknown or only partially specified.

A related limitation concerns how physical information is incorporated into a model. Concatenating physics-derived quantities with observed fields as additional input channels can supply useful descriptors, but it treats physical variables as ordinary features and leaves the network to infer how they should be organized, combined and propagated. Input-level embedding therefore conflates two distinct kinds of physical knowledge—elementary expressions and the governing rules that combine them—and for systems with high-order operators, nonlocal coupling and multiple unresolved constitutive terms, this conflation becomes a structural bottleneck. A model for incomplete physical knowledge should do more than add physics to the input or penalize deviations from a prescribed equation; it should encode the organization of physical knowledge in the computational architecture itself.

Here we introduce the hierarchical physics-embedded adaptive Fourier neural operator (HPE-AFNO), a compositional operator-learning framework for spatiotemporal prediction and constitutive-relation discovery under incomplete physical knowledge. HPE-AFNO represents scientific dynamics through a two-level computational scaffold (Fig. 1a). The first level learns or embeds fundamental physical expressions, such as differential operators, constitutive components or transport descriptors. The second level learns or embeds the governing combinations through which these intermediate quantities determine temporal evolution. Known components are inserted directly through dedicated physics-embedding channels, whereas unknown components are learned by adaptive Fourier neural operator modules. Physical knowledge therefore enters the model as intermediate architectural structure, rather than as a loss-level penalty, a fixed state-update term or an unstructured input augmentation.

This hierarchical formulation provides three advantages, two of which we establish theoretically. First, the factorization admits a hierarchical error decomposition: the overall approximation error is controlled by the first-level extraction errors, the second-level combination error and a compositional residual, so that embedding a known physical component provably removes the corresponding term from the error budget (see 'Hierarchical error decomposition of HPE-AFNO' in Methods). Second, when the hierarchy is aligned with the compositional structure of the dynamics, the number of learnable Fourier parameters sufficient for a prescribed accuracy grows more slowly than for a single-level operator model (see 'Parameter-complexity advantage of hierarchical Fourier approximation' in Methods), thus the benefit of hierarchy is a parameter-complexity property of the architecture rather than a purely empirical observation. Third, the architecture produces intermediate representations that can be inspected and distilled, enabling symbolic recovery of unknown physical expressions within partially specified PDE structures. The adaptive Fourier modules further provide a global spectral inductive bias suited to nonlocal couplings and high-order differential effects, which purely local convolutional updates capture only inefficiently (Fig. 1b).

We evaluate HPE-AFNO across canonical PDE benchmarks and real experimental flow data, covering settings in which physical knowledge is available as known differential operators, partial compositional rules or interpretable transport descriptors. Hierarchical physics embedding improves long-horizon prediction, preserves physically meaningful morphology and spectral structure, remains robust under sparse and noisy observations, and enables the recovery of unknown constitutive relations through symbolic regression. Together, these results establish HPE-AFNO as a unified, theoretically grounded framework for prediction and equation discovery when governing laws are neither fully known nor absent, but partially known and compositionally structured.

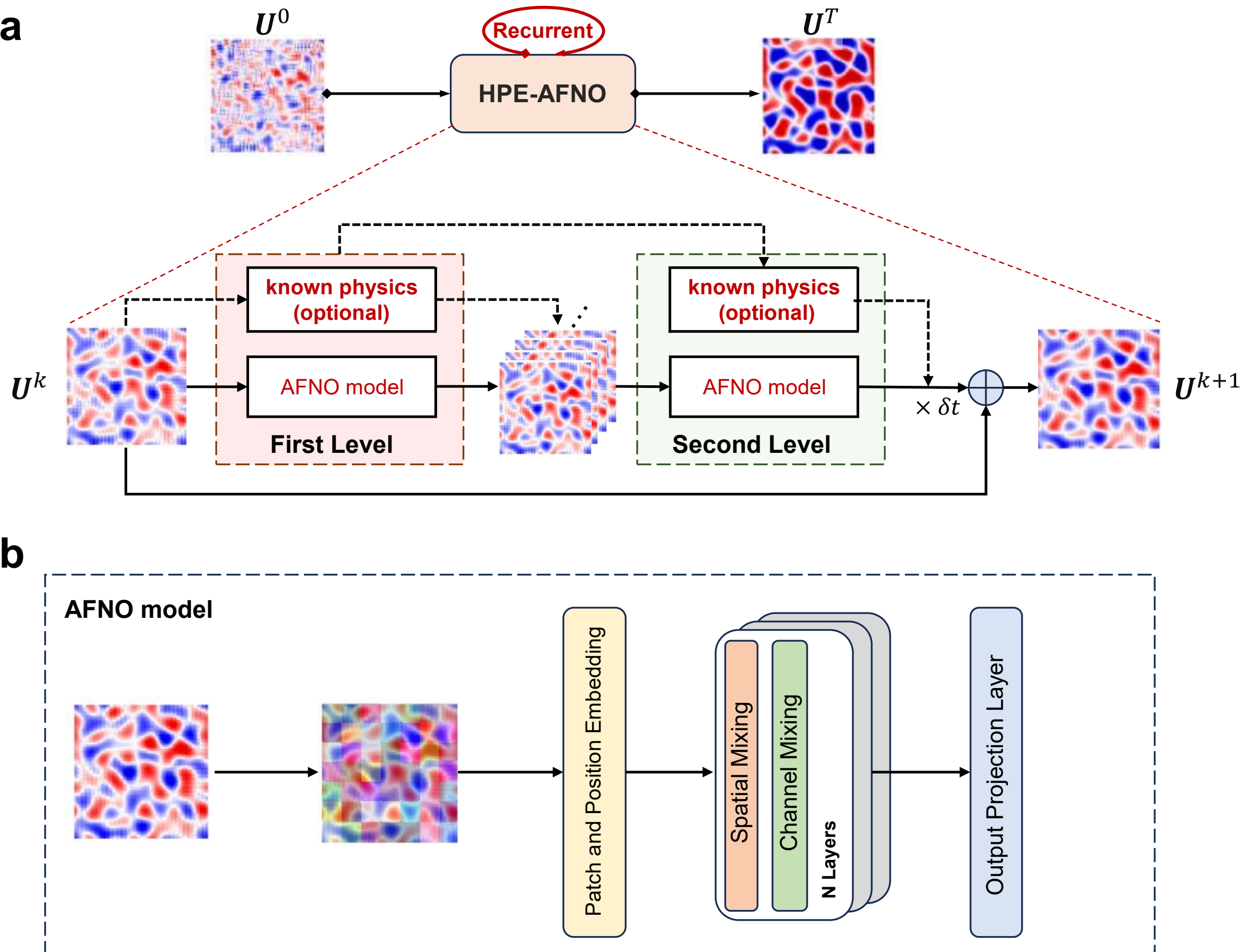

**Fig. 1 | Hierarchical physics-embedded adaptive Fourier neural operator (HPE-AFNO) framework for spatiotemporal dynamics modeling. a**, Schematic of the two-level HPE-AFNO network, illustrating how the model can optionally incorporate known physical knowledge (e.g., diffusivity and free energy terms in phase-field models) and known governing combination rules. Two AFNO modules separately extract fundamental symbolic features from the current field $\boldsymbol{U}^k$ and learn governing combinations among them, culminating in a prediction $\boldsymbol{U}^{k+1}$ via a time-integration step $\delta t$. This hierarchical design not only reduces learning complexity through decomposition, but also enables direct and flexible embedding of known physical knowledge. **b**, Detailed architecture of the AFNO-based feature extraction pipeline. Input fields are first divided into patches with positional information, then processed through multiple spatial and channel mixing layers, ultimately projected back to the physical domain. Fourier transforms in AFNO enhance the model's ability to capture both local and global dynamics.

## 2 Results

### 2.1 Spatiotemporal dynamics prediction under incomplete physical knowledge

We first evaluated hierarchical physics embedding under incomplete specification of the governing dynamics. Many physical systems are known only in part: one may have access to individual differential operators, to selected constitutive components, or to the rule by which physical terms are combined, while the complete closed-form evolution law is unavailable. HPE-AFNO exploits whatever is known by encoding it as intermediate architectural structure, and delegates the remainder—unresolved constitutive terms together with their nonlocal and higher-order couplings— to adaptive Fourier neural operators (see 'HPE-AFNO framework' in Methods). We assessed this capability under limited and noisy observations across systems of increasing complexity, from synthetic phase-field dynamics to experimentally measured wake flows.

We assume that spatiotemporal dynamics are accessible only through a limited number of sparse and potentially noisy snapshots, denoted as $\widetilde{\boldsymbol{U}} \in \mathbb{R}^{N_t' \times C \times \Omega}$, collected with a sampling interval of $\delta t'$. Here, $N_t'$ represents the total number of sparse temporal observations, $C$ denotes the number of state variables or physical channels, and $\Omega$ denotes the spatial grid, which is two- or three-dimensional depending on the system. Our objective is to infer a denoised high-resolution trajectory $\widehat{\boldsymbol{U}} \in \mathbb{R}^{N_t \times C \times \Omega}$, where $N_t' < N_t$, that both reconstructs the dynamics within the observed window ($t \leq t_{N_t'}$) and extrapolates beyond it ($t > t_{N_t'}$) across the temporal horizon.

We began with the Cahn–Hilliard equation[28,29], a canonical phase-field model, to test whether hierarchical physics embedding improves prediction when the governing dynamics are only partially known. The equation describes conserved phase separation and is widely used in materials and electrochemical processes, including the evolution of lithium-ion concentration in intercalation electrodes[30,31]. Its dynamics can be written as

$$\frac{\partial c}{\partial t} = \nabla \cdot \left( M(c) \nabla \left( \mu_{\text{hom}}(c) - \kappa \nabla^2 c \right) \right) \tag{1}$$

where $c(\boldsymbol{x}, t)$ is the conserved concentration field, with $\boldsymbol{x}$ and $t$ denoting the spatial coordinate and time, respectively. The mobility function is defined as $M(c) = cD(c)$, where $D(c)$ is the concentration-dependent diffusivity. The chemical potential consists of a homogeneous bulk contribution, $\mu_{\text{hom}}(c)$, and a gradient-energy contribution, $-\kappa \nabla^2 c$, where $\nabla$ is the spatial gradient operator and $\kappa$ is a positive gradient-energy coefficient. This term penalizes sharp concentration gradients, producing the diffuse interfaces that separate distinct phases—such as lithium-rich and lithium-poor domains—rather than the infinitely sharp boundaries of a purely bulk free energy. Following previous studies, the homogeneous chemical potential and diffusivity are taken as $\mu_{\text{hom}}(c) = \log c(1-c)^{-1} + 3(1-2c)$ and $D(c) = (1-c)$, respectively.

Equation (1) is a concrete instance of a partially specified governing structure, and it exposes the two levels on which HPE-AFNO operates. The first level comprises elementary physical expressions and differential operators—here the Laplacian $\nabla^2 c$, which contributes to the gradient-energy term in the chemical potential. These components correspond to the red-highlighted part of the equation. The second level is the governing combination rule: the prescription by which these elementary quantities are assembled—through the chemical-potential definition, the mobility-weighted chemical-potential gradient, and the outer divergence—into the conservative evolution of the concentration field. This combination rule corresponds to the green-highlighted part of the equation. The compositional structure is therefore not an additional empirical feature to be fitted, but the known law by which physical expressions and operators combine in the governing equation. HPE-AFNO retains this law explicitly and learns only the quantities it cannot specify.

In this experiment, we did not use the available physical knowledge to prescribe the full evolution operator. Instead, HPE-AFNO embedded two complementary pieces of partial knowledge: the local Laplacian operator $\nabla^2 c$ and the governing combination rule of the Cahn–Hilliard dynamics. The remaining unresolved constitutive and nonlinear components were learned from sparse, noisy data. For comparison, PeRCNN[26] and PeSANet[32] were provided the identical Laplacian prior but did not encode the hierarchical rule for assembling physical terms. LNO[33] and FNO[20] served as operator-learning baselines without explicit physics embedding.

As shown in Fig. 2a, at $t = 18.5s$ —well inside the extrapolation regime—HPE-AFNO reproduces the reference concentration morphology more faithfully than any baseline, preserving both the large-scale phase domains and the diffuse interfaces between them. PeRCNN and PeSANet recover the coarse morphology but accumulate visible local distortions, while LNO and FNO over-smooth the field and fail to maintain the high-contrast phase-separated pattern. These visual differences are quantified by the RMSE trajectories in Fig. 2b: HPE-AFNO holds a consistently lower error once prediction crosses from interpolation into extrapolation at $t = 9s$. Averaged over the extrapolation horizon, HPE-AFNO attains an RMSE of 0.052, corresponding to reductions of 39.5%, 44.3%, 72.0% and 65.8% relative to PeRCNN, PeSANet, LNO and FNO, respectively. Because HPE-AFNO and the two physics-encoded baselines share the same Laplacian prior, the gap between them is attributable to the additional structure that HPE-AFNO encodes—the combination rule—which is what stabilizes the long-horizon prediction in Cahn–Hilliard phase separation.

To examine whether the improved field-level accuracy is accompanied by more faithful phase-separation morphology, two physical indicators were further evaluated (Fig. 2c). The characteristic length, estimated from the two-dimensional structure factor, captures the dominant coarsening scale of the phase-separated domains; the radial structure-factor error compares the normalized radial spectrum of the predicted concentration field against the reference, probing whether the spatial organization of the domains is preserved across wavenumbers. HPE-AFNO reduces the characteristic-length error by 54.5% relative to PeRCNN and by 19.9% relative to PeSANet, and reduces the radial structure-factor error by 22.7% and 23.3%, respectively. The model therefore recovers not only the pointwise concentration field but also the coarsening length scale and spectral morphology that define the physics of Cahn–Hilliard separation.

The improvement in the Cahn–Hilliard system follows from how the learning problem is factorized rather than from any single embedded term. Fixing the Laplacian and the combination rule removes them from the set of quantities that need to be inferred: the network is left to estimate only the residual closure—the constitutive nonlinearities and their couplings—within a structure that already enforces conservation and the correct differential order. This shrinks the effective hypothesis space and, under sparse and noisy observations, lowers the variance of the estimator, which is precisely the regime in which long predictions are difficult. The contrast with the baselines is instructive. PeRCNN and PeSANet are provided the same Laplacian prior but treat it as a local correction term; lacking the combination rule, they still need to learn how the physical quantities are assembled into a conservative update, and small errors in that assembly compound over the horizon. LNO and FNO, with no physics embedded, must learn the entire evolution operator from the same scarce data. We next tested whether the hierarchical physics embedding remains effective in a more demanding three-dimensional setting, where volumetric interface motion, stronger spatial coupling and accumulated extrapolation error make long-horizon prediction substantially harder.

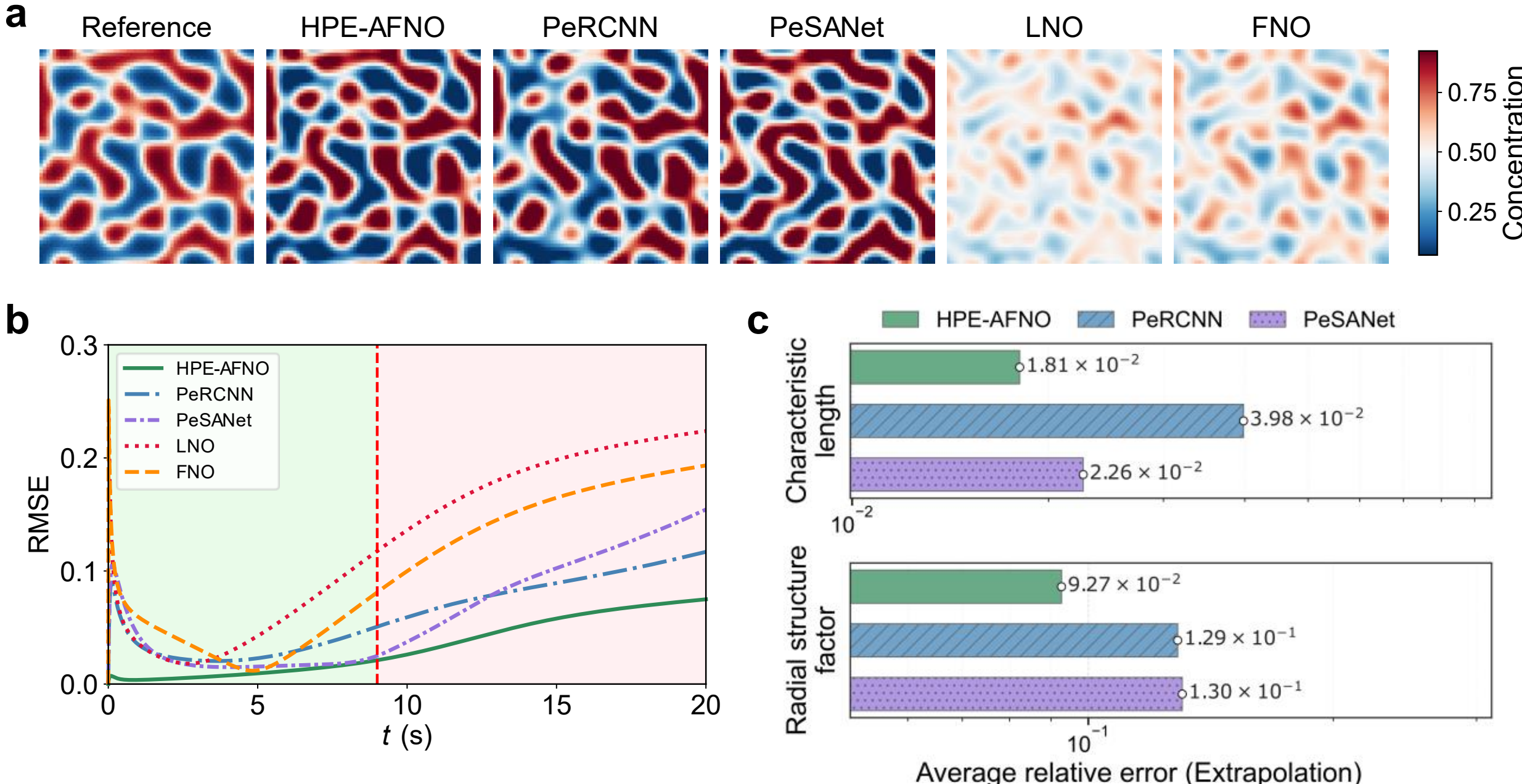

**Fig. 2 | Comparative performance of HPE-AFNO and baseline models on Cahn–Hilliard phase-separation dynamics. a**, Predicted concentration fields at $t = 18.5s$, within the extrapolation regime, comparing the reference against HPE-AFNO, PeRCNN, PeSANet, LNO and FNO. The color scale denotes the normalized, dimensionless concentration. **b**, RMSE trajectories over the prediction horizon; the red dashed line at $t = 9s$ separates the interpolation regime (green) from the extrapolation regime (pink). **c**, Relative errors in the characteristic length and the radial structure-factor, averaged over the extrapolation interval ($t > 9s$). The characteristic length captures the dominant coarsening scale of the phase-separated domains, while the radial structure-factor error quantifies how faithfully the

predicted field reproduces the reference spectral morphology across radial wavenumbers. Errors were computed between predicted and reference rollouts at matched time points. HPE-AFNO achieves lower pointwise prediction error and better preserves phase-separation morphology, with its advantage widening under extrapolation.

The three-dimensional Allen–Cahn equation[13,14] describes non-conserved phase-field dynamics driven by interfacial smoothing and local double-well relaxation. In this work, the dynamics are given by

$$\frac{\partial c}{\partial t} = \nabla^2 c - (c^3 - c) \tag{2}$$

where $c(\boldsymbol{x}, t)$ is the phase-field variable and $\boldsymbol{x} = (x, y, z)$ denotes the three-dimensional spatial coordinate. The Laplacian term $\nabla^2 c$ represents interfacial smoothing, whereas the nonlinear reaction term $-(c^3 - c)$ drives the local relaxation of the order parameter towards the two stable phases. In this experiment, the available physical knowledge was restricted to the Laplacian operator. HPE-AFNO, PeRCNN and PeSANet therefore received the same local differential prior; the nonlinear reaction term was withheld and learned from data. This setup is a stricter test than the Cahn–Hilliard case, because the only embedded structure is now shared by all physics-informed models.

As shown in Fig. 3a, HPE-AFNO more accurately tracks the three-dimensional Allen–Cahn evolution across the prediction horizon. It preserves the volumetric structures and their progression from the initial condition into the extrapolation regime. PeRCNN and PeSANet reproduce the coarse domain morphology but show increasingly visible deviations in local interfacial structures, whereas FNO and LNO lose the volumetric evolution more rapidly. These observations are consistent with the RMSE trajectories: HPE-AFNO maintains the lowest extrapolation error after $t = 9s$ (Fig. 3b). Beyond pointwise accuracy, two physical indicators assess whether the predicted dynamics remain thermodynamically consistent (Fig. 3c). The free-energy error measures how far the predicted energetic relaxation departs from the reference trajectory—that is, whether the model follows the correct thermodynamic driving force—while the bulk-to-interface energy ratio measures whether the predicted field maintains the proper balance between bulk relaxation and interface formation. HPE-AFNO attains the lowest averaged extrapolation error on both, showing that its advantage is not limited to pointwise field prediction but also extends to the preservation of physically meaningful energetic structure that governs its evolution.

These results rule out a simpler interpretation that the improvement arises merely from including a known Laplacian. The same local differential prior is also provided to PeRCNN and PeSANet. In HPE-AFNO, however, the Laplacian is represented as an intermediate physical expression within a hierarchical operator-learning structure, where it is further coupled with learned latent representations through global spectral mixing. This allows the model to use the local interfacial-smoothing prior while also capturing nonlocal correlations among evolving three-dimensional interfaces. Such interactions are particularly important in volumetric phase-field dynamics, where long-horizon errors can accumulate through distortions of global morphology and energetic balance, not only through local interface motion. Therefore, HPE-AFNO provides a more effective mechanism for transforming the same embedded physical operator into physically consistent long-horizon predictions.

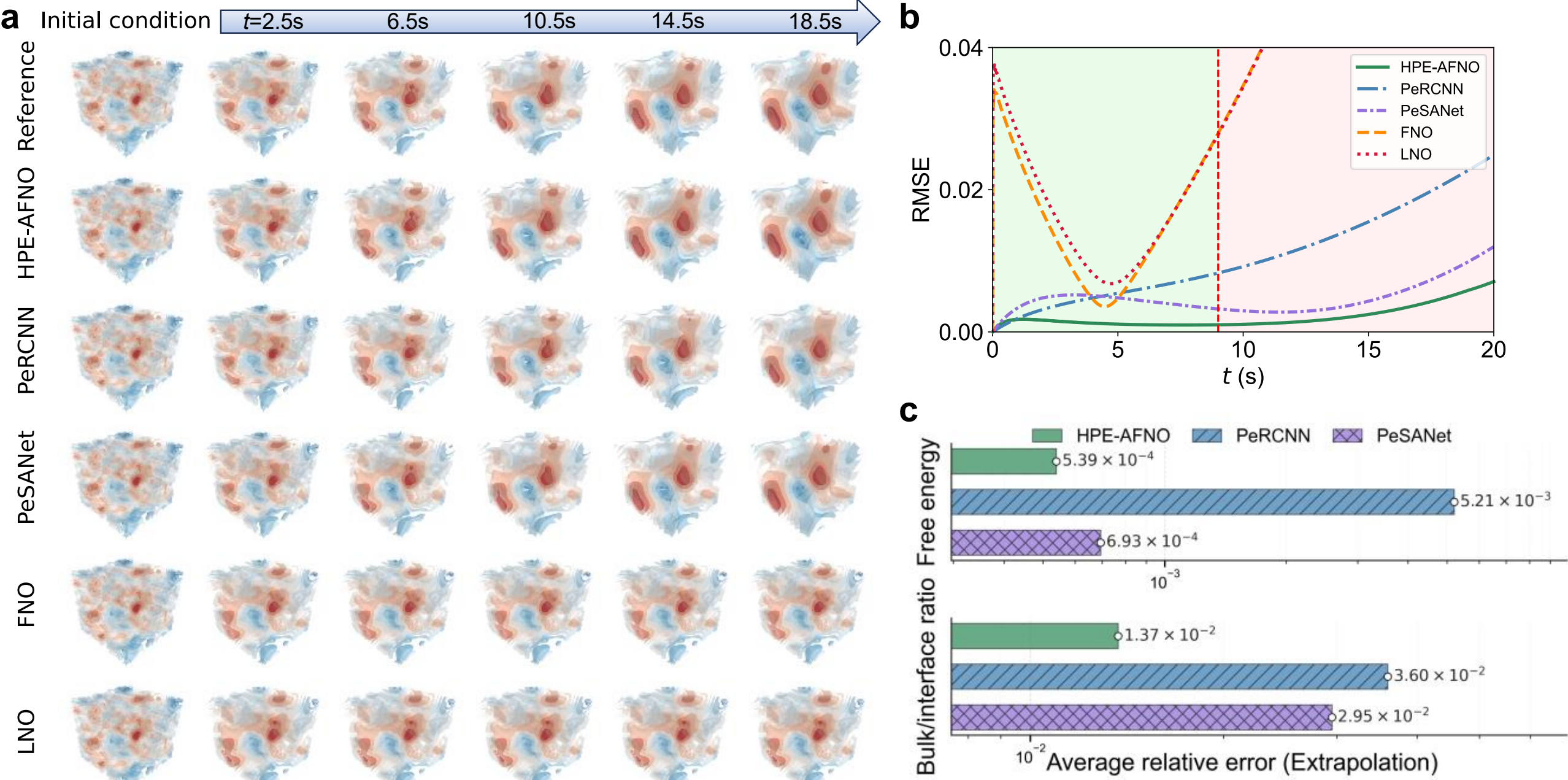

**Fig. 3 | Comparative performance of HPE-AFNO and baseline models on three-dimensional Allen–Cahn dynamics. a**, Predicted evolution of the three-dimensional Allen–Cahn system at representative time points from the initial condition to $t = 18.5s$, comparing the reference against HPE-AFNO, PeRCNN, PeSANet, FNO and LNO. **b**, RMSE trajectories over the prediction horizon; the dashed line at $t = 9s$ separates the interpolation regime (green) from the extrapolation regime (pink). **c**, Relative errors in the free energy and the bulk-to-interface energy ratio, averaged over the extrapolation interval $(t > 9s)$. The free-energy error measures whether the predicted dynamics reproduce the energetic relaxation of the reference system, while the bulk-to-interface ratio error measures whether the balance between bulk free-energy and interfacial contributions is preserved. Errors were computed between predicted and reference values at matched time points. HPE-AFNO attains lower error and stronger energetic consistency in three-dimensional phase-field prediction.

Synthetic PDEs allow controlled evaluation, but experimental spatiotemporal data introduce additional sources of error, including measurement noise, boundary imperfections, limited spatial resolution and unresolved three-dimensional effects. To assess practical performance under these non-ideal conditions, we applied HPE-AFNO to the experimental hydrofoil-wake dataset from RealPDEBench[34]. The dataset comprises spatiotemporal velocity fields around a hydrofoil and provides a benchmark for fluid dynamics prediction with coherent vortex shedding and downstream wake evolution.

Because no closed-form equation can be trusted for these measurements, we provided partial fluid-mechanical knowledge in the form of physically interpretable transport descriptors computed directly from the measured two-dimensional velocity field $\boldsymbol{u} = (u, v)$. Rather than impose an idealized governing equation that the noise, unresolved three-dimensional effects and other uncharacterized disturbances would violate, we selected local transport quantities directly relevant to wake evolution. Specifically, the vorticity and nonlinear convective descriptors are computed as

$$\omega = \frac{\partial v}{\partial x} - \frac{\partial u}{\partial y}, A_x = u\frac{\partial u}{\partial x} + v\frac{\partial u}{\partial y}, A_y = u\frac{\partial v}{\partial x} + v\frac{\partial v}{\partial y} \quad (3)$$

Here, $\omega$ characterizes rotational structures and vortex shedding in the hydrofoil wake, whereas $A_x$ and $A_y$ represent the two components of convective acceleration associated with nonlinear momentum transport. These quantities provide partial physical information without imposing a closed-form fluid equation. In HPE-AFNO, the computed vorticity and convective descriptors are introduced through dedicated physics-embedding channels and combined with data-driven representations learned from the raw velocity fields. This design allows the model to use physically interpretable transport descriptors as structured guidance while retaining sufficient flexibility to learn unresolved experimental effects from data. PeRCNN and PeSANet were not supplied with explicit physical terms in this real-data experiment. Their physics-encoding mechanisms are better suited to cases where the available physics can be written as local differential or algebraic terms that directly enter the state-update equation, as in reaction–diffusion or phase-field systems. In the hydrofoil-wake measurements, the computed vorticity and convective descriptors provide useful information about rotational motion and nonlinear transport, but they do not by themselves define the temporal update of the measured velocity components. Their effects are entangled with pressure, viscous, boundary and unresolved three-dimensional contributions. These descriptors were therefore used as intermediate physical representations in HPE-AFNO rather than as equation-level update terms in PeRCNN or PeSANet.

The predictive performance on the real experimental hydrofoil wake is summarized in Fig. 4. With vorticity and convective descriptors embedded as intermediate representations, HPE-AFNO reconstructs the temporal evolution of both velocity components most accurately (Fig. 4ab). For the streamwise velocity $u$, every model captures the dominant wake, but HPE-AFNO possesses smaller signed errors in the downstream region and better preserves the spatial organization of the flow. For the transverse velocity $v$, where vortex-induced fluctuations are more sensitive to measurement noise and long-horizon error accumulation, the advantage of HPE-AFNO becomes more evident: the baseline models show stronger smoothing, phase distortion or weakened fluctuation patterns, whereas HPE-AFNO retains more coherent wake evolution over time. The distributions of spatial RMSE per time step confirm this trend (Fig. 4c). For the $u$ component, HPE-AFNO reduces the median spatial RMSE per time step by 34.2% and 30.4% relative to PeRCNN and PeSANet, respectively. For the $v$ component, the corresponding reductions are 53.0% and 50.2%, respectively.

We also evaluated whether the predicted fields preserve flow structure beyond pointwise velocity accuracy. The time-averaged kinetic-energy spectrum (Fig. 4d) shows that HPE-AFNO more closely follows the reference spectrum over the resolved wavenumber range, particularly at higher wavenumbers associated with finer wake structures. Vorticity fields computed from the predictions (Fig. 4e) further show that HPE-AFNO retains more coherent downstream vortices, whereas baseline predictions are more diffuse or less organized. These results indicate that embedding interpretable transport descriptors as intermediate physical representations improves both velocity-field prediction and the preservation of spectral and vortical wake structure in real experimental data.

Across these three systems, the results show that HPE-AFNO improves spatiotemporal prediction across increasingly challenging forms of incomplete physical knowledge, from synthetic phase-field systems with explicit operators to real experimental flow data with only physically interpretable transport descriptors. This hierarchical separation between prescribed physical components and learned unresolved dynamics also yields intermediate representations that can be interrogated for scientific discovery. We therefore next examine whether these representations, combined with symbolic regression, can recover unknown constitutive relations in partially specified PDEs from sparse and noisy data.

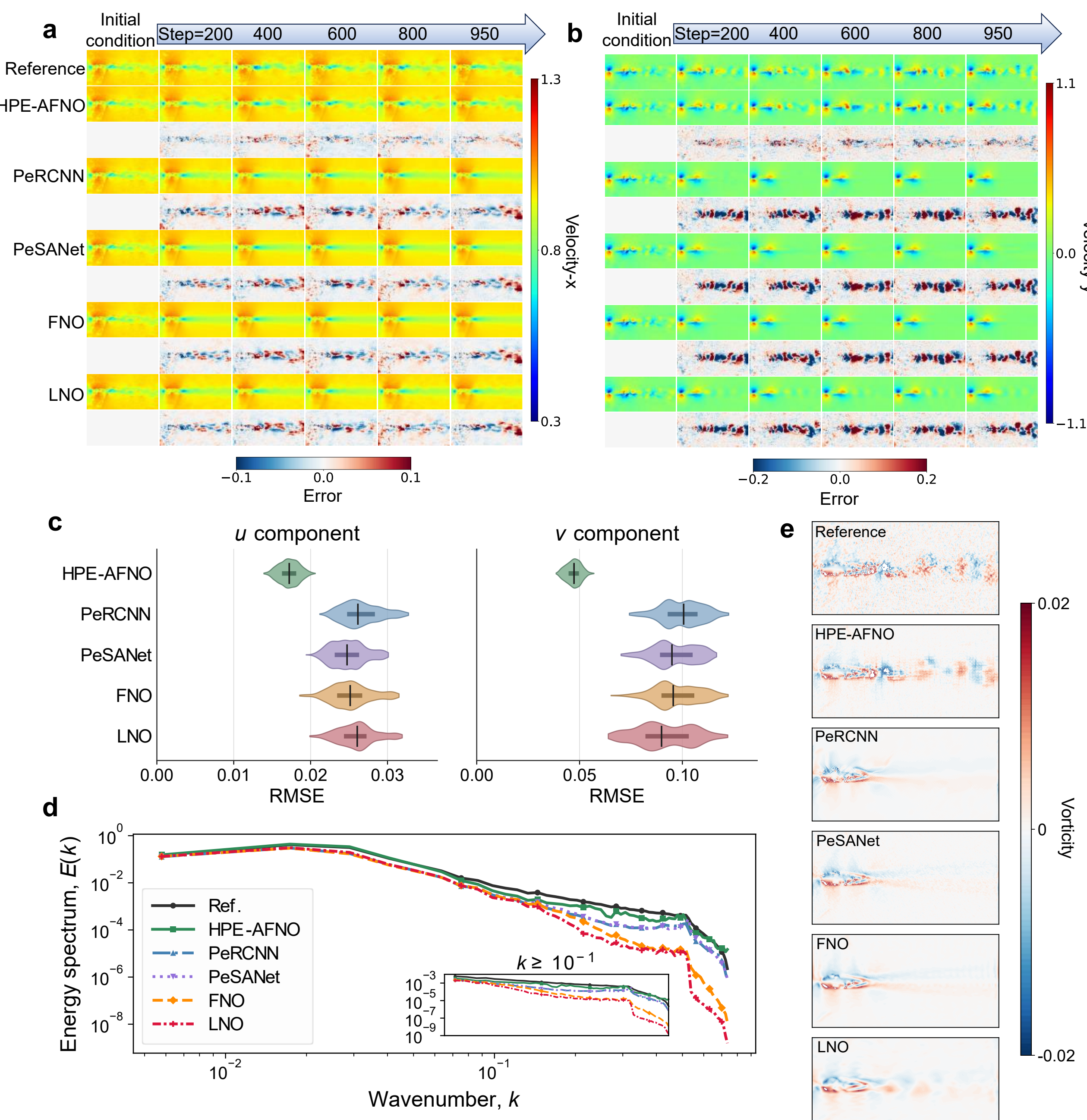

**Fig. 4 | Comparative predictive performance on real experimental hydrofoil-wake dynamics. a, b,** Predicted evolution of the streamwise velocity $u$ (**a**) and transverse velocity $v$ (**b**)from step 200 to step 950. For each component, the reference is compared against HPE-AFNO, PeRCNN, PeSANet, FNO and LNO, with signed prediction-error maps shown beneath each prediction. Color bars report field values and signed errors on the evaluation-data scale; no physical units are assigned. **c**, RMSE distributions for the $u$ and $v$ components, computed over the spatial grid at each rollout step; violin plots summarize the temporal distribution across the prediction horizon. **d**, Time-averaged kinetic-energy spectrum, obtained from the two-dimensional Fourier transform of each velocity field after removing its spatial mean, showing how fluctuation energy is distributed across wavenumbers and whether fine-scale wake structures are retained. **e**, Vorticity fields derived from the predicted velocities at step 600, assessing preservation of rotational structures and vortex-shedding patterns in the experimental flow.

## 2.2 Discovery of unknown physical expressions from limited and noisy data

Forward prediction is only one component of scientific machine learning. In many applications, the aim is also to extract interpretable physical relationships from data. We therefore evaluate whether HPE-AFNO can support constitutive-relation discovery when the overall PDE structure is partially known but selected physical expressions remain unspecified.

We considered the nonlinear Cahn–Hilliard system in equation (1). Rather than aiming to discover the complete governing PDE from scratch, we considered a partially specified setting in which the general equation structure is known, but selected physical expressions remain unknown. HPE-AFNO is integrated with deep symbolic regression (DSR)[35] to identify these unknown expressions. As illustrated in Fig. 5, the workflow consists of three steps: (1) reconstruct the spatiotemporal dynamics from sparse, noisy observations with HPE-AFNO; (2) apply concentration binning to extract the functional dependence of each learned unknown term on the concentration field; and (3) use DSR to recover a compact analytical expression for each term. Details are provided in the Methods section.

We validated the framework on synthetic Cahn–Hilliard data under sparse, noisy measurement conditions. The general form of the Cahn–Hilliard equation (equation (1)) was treated as known, while the symbolic forms of the concentration-dependent diffusivity $D(c)$ and the homogeneous chemical potential $\mu_{\mathrm{hom}}(c)$ were withheld as unknowns. The known components were embedded into HPE-AFNO to guide the reconstruction of the concentration dynamics (Fig. 5). To make the learned unknown terms interpretable as functions of concentration, we introduced a kernel-based mapping that enforces continuous dependence on the input concentration field, so that locations with similar concentration values yield similar encoded representations. After reconstruction, concentration binning was utilized to provide empirical functional relationships for the learned $D(c)$ and $\mu_{\mathrm{hom}}(c)$, which were then passed to DSR without prescribing their analytical forms. The recovered expressions were assessed against both binned ground-truth relationships and the final symbolic forms.

The results show that HPE-AFNO accurately recovered both unknown expressions from sparse and noisy measurements. Even when the noise level was increased to 10% (Supplementary Note 4 and Supplementary Figs. 1 and 2), our approach maintained accurate identification of both analytical expressions for the diffusivity $D(c)$ and homogeneous chemical potential $\mu_{\mathrm{hom}}(c)$. In contrast, PeRCNN followed by sparse regression, as well as end-to-end DSR and genetic programming baselines, failed to identify the mathematical forms of either $D(c)$ or $\mu_{\mathrm{hom}}(c)$ under the same conditions (Supplementary Note 5). These comparisons show that the hierarchical reconstruction-and-discovery strategy is especially beneficial when temporal observations are sparse and contaminated by noise.

We further assessed the discovery framework on the three-dimensional Allen–Cahn equation by withholding the nonlinear reaction term. The reconstructed three-dimensional fields closely match the reference dynamics at representative time points, providing a reliable basis for downstream symbolic analysis (Extended Data Fig. 1a). The learned reaction representation recovers the expected cubic dependence on the state variable and closely follows the reference term $u - u^3$, with a low reconstruction error (Extended Data Fig. 1b). The reward–complexity analysis identifies $u - u^3$ as a compact, high-reward and physically interpretable expression among the candidate symbolic forms (Extended Data Fig. 1c). The results indicate that the framework extends beyond the Cahn–Hilliard case to another class of phase-field dynamics.

The discovery advantage arises from two features of the framework. First, embedding the known PDE structure enables denoised reconstruction of spatiotemporal trajectories before symbolic analysis, reducing reliance on direct temporal-derivative estimation from sparse and noisy measurements. This is important because end-to-end symbolic regression is typically sensitive to derivative errors when temporal resolution is limited. Second, the hierarchical architecture assigns different unknown physical components to separate representation channels, so each term can be analyzed and fitted individually rather than being recovered as part of a single entangled residual. This separation reduces the difficulty of identifying multiple coupled terms and produces intermediate representations that are more suitable for symbolic regression.

Methods that reconstruct the dynamics and then apply sparse regression to a global candidate library remain more vulnerable to reconstruction and derivative-estimation errors. They also depend on predefined libraries of candidate functions, which can fail when the correct non-polynomial or compositional form is absent (Supplementary Note 5). In contrast, the HPE-AFNO-DSR workflow uses partial structural knowledge to stabilize reconstruction, separates unknown physical expressions into interpretable channels and then searches for compact symbolic forms. These results show that HPE-AFNO provides an effective route for constitutive-relation discovery in partially specified PDE systems, particularly under sparse and noisy observations.

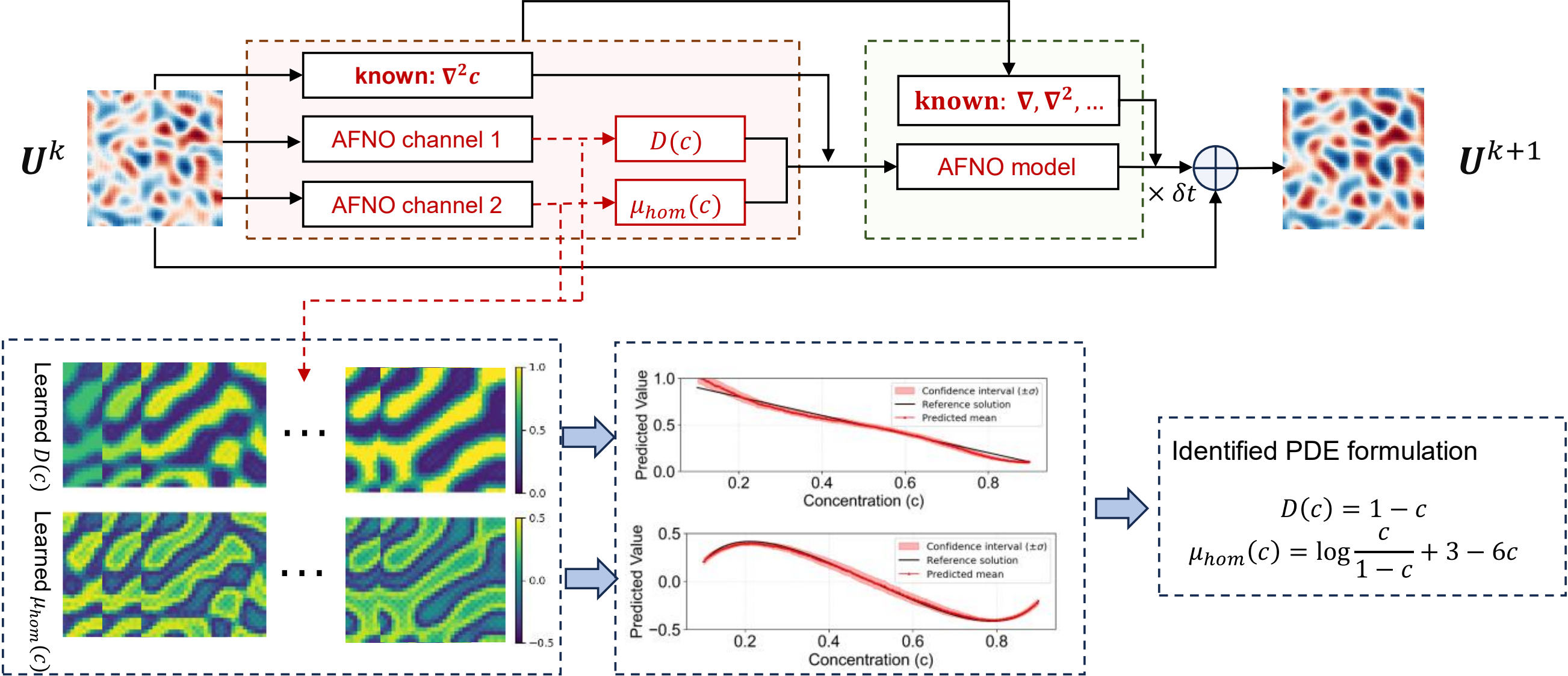

**Fig. 5 | Framework for discovering unknown physical expressions using hierarchical physics-embedded learning.** The procedure begins with the reconstruction of spatiotemporal dynamics from sparse and noisy measurements using the HPE-AFNO model. In the first level, two separate AFNO channels are used to learn the unknown physical terms: the concentration-dependent diffusivity $D(c)$ and the homogeneous chemical potential $\mu_{hom}(c)$, respectively. The second level embeds known physical operators, such as $\nabla^2$, to produce the next-step system's state $U^{k+1}$. To extract interpretable physical relationships, the predicted concentration field is subjected to binning analysis, which reveals the underlying functional dependencies without assuming specific forms. Finally, deep symbolic regression is applied to recover closed-form expressions of the unknown terms. The workflow is illustrated using the partially specified Cahn–Hilliard equation as an example.

## 3 Discussion

Scientific machine learning often works between two idealized extremes: a closed governing equation that can be solved or constrained directly, and a purely empirical process for which only black-box prediction is possible. The systems considered here occupy the more common intermediate regime. They possess partial physical structure—differential operators, conservation forms or transport descriptors—but the constitutive relations, closures or experimental effects required to complete the dynamics are not fully specified. This distinction is important. When physical knowledge is incomplete, enforcing an assumed equation can introduce model bias, whereas ignoring available structure increases sample complexity and weakens extrapolation. HPE-AFNO addresses this regime by representing partial physical knowledge as a compositional computational scaffold rather than as an external penalty, a fixed update rule or an unstructured auxiliary input.

The central mechanism is the separation of physical expressions from their governing combinations. In classical scientific modeling, equations are rarely monolithic objects: they are assembled from elementary operators, constitutive functions and combination rules that encode conservation, transport or energetic relaxation. HPE-AFNO mirrors this organization architecturally.

Known components can be embedded at the level at which they are physically meaningful, while unresolved terms remain learnable through AFNO. This decomposition reduces the burden on the data-driven component: the network does not need to rediscover all operators, couplings and assembly rules simultaneously, but only the missing parts of a partially specified dynamical structure.

This perspective helps explain the robustness observed under sparse and noisy observations. The contour plots in Extended Data Fig. 2 should not be interpreted merely as a performance summary; they reveal how the model behaves as the information content of the data is degraded. The smooth increase in error with sampling interval and noise level shows that hierarchical embedding does not create a brittle dependence on finely resolved measurements. Instead, the embedded physical scaffold constrains the range of admissible dynamics, while the recursive rollout and spectral mixing provide a mechanism for interpolating unresolved intermediate states and propagating long-range spatial correlations. In phase-field systems, this matters because morphology evolves through coupled local interface motion and global coarsening. A model that preserves only local smoothness can still drift in domain scale, spectral organization or energetic balance. By retaining operators and combination structure as intermediate representations, HPE-AFNO regularizes the trajectory at the level of the dynamics rather than only at the level of individual predicted fields.

The comparison with single-level physics embedding further clarifies what kind of prior is useful. Directly concatenating vorticity and convective descriptors with the velocity field provides the network access to the same physical variables, but not to their role in the dynamics. In that setting, the descriptors become ordinary channels, and the model must infer from data how rotational motion, nonlinear transport and unresolved effects should interact during rollout. HPE-AFNO instead preserves these quantities as structured intermediate representations before integrating them with learned spatiotemporal features. The improved wake prediction in Extended Data Fig. 3 therefore supports a mechanistic conclusion: the value of physics embedding lies not only in what physical information is provided, but in how that information is organized. For experimental flow data, where pressure, boundary effects, viscous dissipation, three-dimensional disturbances and measurement noise are entangled, such organization is more appropriate than imposing an idealized closed equation or treating all descriptors as exchangeable features.

This finding positions HPE-AFNO between several established scientific machine-learning paradigms. Physics-informed neural networks are powerful when governing equations are known with sufficient fidelity, because the residual of the equation can act as a soft constraint during training. Their strength becomes a limitation when essential terms are missing or when the measured system deviates from the assumed equation. Neural operators, by contrast, learn mappings between function spaces and can provide efficient surrogates for complex PDE dynamics, but their physical structure is largely implicit; without additional organization, they often require more data and accumulate nonphysical drift in long rollouts. Physics-encoded recurrent or convolutional models hard-wire selected local operators into state updates, which improves physical consistency when the update form is known, but they typically do not represent the compositional hierarchy by which physical expressions are assembled. HPE-AFNO complements these approaches by treating partial knowledge as structured, modular and incomplete: known physical fragments guide learning without requiring the full equation to be specified.

The discovery results extend this argument from prediction to interpretation. A key difficulty in equation discovery is that sparse and noisy measurements make derivative estimation unstable, while global sparse-regression libraries can entangle multiple unknown terms into a single residual. HPE-AFNO changes the discovery problem by first reconstructing a denoised trajectory under partial structural constraints and then exposing intermediate learned components that correspond to candidate physical expressions. Symbolic regression is therefore applied not to an unconstrained residual, but to separated representations that already inherit the organization of the partially known PDE. This is why the framework is able to recover constitutive relations under conditions where end-to-end symbolic search or sparse regression fails. The broader implication is that interpretability in scientific machine learning need not be achieved solely after training; it can be designed into the representational structure of the model.

The approach also shows a more nuanced view of "physics-informed" learning. Physical knowledge is often treated as either a complete law to be enforced or a collection of features to be appended. Many real systems require an intermediate treatment. A conservation form may be trusted while the closure is not; a differential operator may be known while its coefficient is state-dependent and unknown; transport descriptors may be mechanistically meaningful but insufficient to define an update equation. HPE-AFNO provides a way to encode such partial knowledge. It allows one to specify which components are reliable, which should remain learnable and how they should be combined. This is especially relevant for experimental scientific data, where the gap between idealized equations and measured dynamics is often produced by unresolved dimensions, imperfect boundaries, heterogeneous material properties and sensor noise.

## 4 Conclusions

This work presents HPE-AFNO as a compositional physics-embedded learning framework for spatiotemporal systems governed by incomplete physical knowledge. The main contribution is not simply the addition of physics-derived quantities to a neural operator, but the architectural organization of partial physical knowledge into hierarchical representations. Fundamental physical expressions are embedded or learned as intermediate components, and their governing combinations are then assembled to determine temporal evolution. This design reflects the way scientific models are constructed: from operators, constitutive relations and coupling rules, some of which are known and others of which must be inferred. Across phase-field systems and experimental wake dynamics, the framework improves long-horizon prediction while preserving physically meaningful morphology, energetic relaxation, spectral organization and vortical structures. The ablation against single-level physics embedding shows that these improvements arise from structured use of physical information, not merely from additional input channels. The robustness analysis further indicates that hierarchical embedding remains useful when observations are temporally sparse and contaminated by noise, a regime in which purely data-driven operator learning and derivative-dependent methods are especially vulnerable.

Beyond prediction, HPE-AFNO provides a route to constitutive-relation discovery under partial structural knowledge. By separating known and unknown physical components into interpretable intermediate channels, the model produces representations that can be distilled into compact symbolic expressions. This bridges two objectives that are often treated separately in scientific machine learning: accurate forecasting and mechanistic insight. The resulting framework does not require a fully specified PDE, but also does not discard physical knowledge when it is incomplete. These findings suggest that scientific machine learning need not choose between purely data-driven neural operators and fully prescribed physics-based models. A more flexible alternative is to encode partial physical knowledge as a compositional scaffold, within which unresolved mechanisms remain learnable and interpretable. Future work should extend this principle to systems with stronger multi-physics coupling, irregular geometries, imperfect or conflicting physical priors, and greater experimental heterogeneity. As scientific applications increasingly involve incomplete governing laws, sparse measurements and unresolved constitutive relations, hierarchical physics-embedded learning provides a practical route for modeling spatiotemporal dynamics in a way that remains both predictive and interpretable.

# 5 Methods

## 5.1 HPE-AFNO framework

Consider a spatiotemporal dynamical system whose spatiotemporal evolution is governed by a partial differential equation (PDE) with high-order differential terms and nonlinear operators:

$$\boldsymbol{u}_t = \mathcal{F}(\boldsymbol{u}, \boldsymbol{u}_x, \boldsymbol{u}_{xx}, \boldsymbol{u}_{xxx}, \boldsymbol{u}_{xxxx}, \dots, \boldsymbol{x}, t) \tag{4}$$

where $\boldsymbol{u}(\boldsymbol{x}, t) \in \mathbb{R}^n$ represents the state variable—namely, the observational data or snapshots collected from experiments or nature—with $(\boldsymbol{x}, t)$ ranging over the space–time domain $\Omega \times T$. $\boldsymbol{u}_{\boldsymbol{x}}$ and $\boldsymbol{u}_{\boldsymbol{xx}}$ denotes the first-order and second-order spatial derivatives with respect to the spatial variable $\boldsymbol{x}$, respectively, and the same convention applies to higher-order derivatives; $\boldsymbol{u}_t$ represents the first-order time derivative; and $\mathcal{F}$ is a nonlinear function involving $\boldsymbol{u}$ and its spatial derivatives of various orders with nonlinear operators such as logarithmic functions. The solution is specified together with $\mathcal{I}(\boldsymbol{u}; t = 0, \boldsymbol{x} \in \Omega) = 0$ and $\mathcal{B}(\boldsymbol{u}; \boldsymbol{x} \in \partial\Omega) = 0$ as the initial and boundary conditions, respectively.

Through numerical discretization with time step $\delta t$, we reformulate this PDE using a forward Euler scheme, enabling the construction of a spatiotemporal learning model where the state variable $\boldsymbol{u}$ is updated via a recurrent network with shared parameters:

$$\hat{\boldsymbol{u}}^{k+1} = \hat{\boldsymbol{u}}^k + \hat{\mathcal{F}}(\hat{\boldsymbol{u}}^k, \theta)\delta t \tag{5}$$

where, $\hat{\boldsymbol{u}}^k$ represents the prediction at time $t_k$, and $\hat{\mathcal{F}}$ is an approximation of $\mathcal{F}$ parameterized by $\theta$. The approximation functions as a proxy for a series of operations responsible for calculating the first-order time derivative as specified in equation (4).

A key challenge in this framework is developing an autoregressive neural network capable of accurately learning $\mathcal{F}$ through frame-by-frame prediction, ensuring robust recursive updates of the state variable $\boldsymbol{U}^k \in \mathbb{R}^{H \times W}$ (in this work, we mainly study two-dimensional cases). We introduce a hierarchical architecture that mirrors the progressive discovery process of physical laws (as shown in Fig. 1). Specifically, we implement a two-level architecture: the first level extracts fundamental physical terms, and the second level learns the governing interactions and combinations among these fundamental terms. Take the CH equation as an illustrative example, where the state variable is the concentration $c$ (i.e. $u = c$ ). The first level specializes in extracting spatiotemporal features corresponding to constitutive relations with respect to $c$ (i.e., the product of concentration and concentration-dependent diffusivity $M(c)$, the homogeneous chemical potential $\mu_{\text{hom}}(c)$, and the gradient energy term $-\kappa\nabla^2 c$, where $\nabla$ denotes the spatial gradient). To achieve this, we employ an attention layer based on a Fourier mixer that leverages the powerful expressiveness of vision transformers. The Fourier layer $\boldsymbol{Y}^k = \text{FourierLayer}(\boldsymbol{U}^k)$ consists of three parts: (1) employ the discrete Fourier transform to mix different tokens: $\boldsymbol{Z}^k[i,j,:] = \text{DFT}(\overline{\boldsymbol{U}}^k)[i,j,:]$ where $\overline{\boldsymbol{U}}^k \in \mathbb{R}^{H' \times W' \times d}$ is obtained through patch and position embedding from the input $\boldsymbol{U}^k$; (2) multiply the weight matrix $\boldsymbol{W}_f \in \mathbb{C}^{H' \times W' \times d \times d}$ in the frequency domain to mix different channels: $\overline{\boldsymbol{Z}}^k[i,j,:] = \boldsymbol{W}_f[i,j,:,:]\boldsymbol{Z}^k[i,j,:]$; and (3) adopt the inverse discrete Fourier transform: $\boldsymbol{Y}^k[i,j] = \text{IDFT}(\overline{\boldsymbol{Z}}^k)[i,j]$. To further reduce the number of learnable parameters and improve the adaptability, the multilayer perceptron (MLP) with shared parameters is adopted in step (2): $\bar{\boldsymbol{Z}}^k[i,j,:] = \boldsymbol{W}_{f2}\sigma\left(\boldsymbol{W}_{f1}\boldsymbol{Z}^k[i,j,:]\right) + \boldsymbol{b}$ [20,21]. This layer excels at capturing global dependencies between inputs and outputs through frequency-domain learning, especially in modeling complex nonlinear (e.g., non-polynomial) operators compared to CNN-based methods that rely on local convolutional kernels. For multi-feature outputs, we modify the first-level AFNO module to generate multiple channels that feed into the second-level combination learning. The second level learns the complex interactions and combinations between $M(c)$, $\mu_{\text{hom}}(c)$ and $-\kappa\nabla^2 c$ (i.e., $\nabla\cdot\left(M(c)\nabla(\mu_{\text{hom}}(c)-\kappa\nabla^2 c)\right)$). We utilize the same Fourier attention layer as the first-level. This hierarchical design significantly reduces learning complexity through decomposition.

Moreover, it enables more direct and flexible embedding of known physical constraints compared to previous approaches like PeRCNN. For a known fundamental physical term, we can create a separate embedding channel to directly compute and fuse it with other outputs at the first level. For example, if we know $\mu_{\text{hom}}(c) = \log c(1-c)^{-1} + 3(1-2c)$, we can directly calculate this term in a separate channel and feed it with the outputs of the Fourier attention channels (corresponding to the unknown terms $M(c)$ and $-\kappa\nabla^2 c$) into the second-level learning. For known governing interactions and combinations, we utilize the outputs of the Fourier attention channels in the first-level as the corresponding unknown physical terms, and the results of the separate embedding channels as the corresponding known terms. With these fundamental terms, we can form a separate embedding channel in the second-level to compute the results following these known combinations. This structured embedding enforces physical consistency in predictions, substantially enhancing both predictive accuracy for long-term dynamics and physical interpretability. This behavior is demonstrated by the numerical experiments. In summary, physics-embedding channels enforce known physical laws, while Fourier attention channels model the residual (potentially complex) dynamics at both levels.

### 5.2 Hierarchical error decomposition of HPE-AFNO

The purpose of the HPE-AFNO hierarchy is not only to improve expressivity, but also to reduce the difficulty of learning a complex PDE evolution operator, especially when partial physical knowledge is available. Rather than approximating the full operator in a single stage, HPE-AFNO decomposes the task into two coupled subproblems: extracting a set of fundamental physical components from the current state, and learning how these components are combined into the governing dynamics. This decomposition is especially well suited to scientific systems whose dynamics are generated by nested compositions of constitutive components, differential operators and nonlinear couplings. In such settings, a single-level approximation needs to absorb multiple sources of complexity within one mapping, whereas a hierarchical approximation can distribute them across levels and directly replace known components or known combination rules by embedded physics channels. This mechanism motivates the following error decomposition for the proposed architecture.

To formalize this point, we assume that the target evolution operator $\mathcal{F}$ admits an approximately compositional representation:

**Assumption 1.** *There exists an ideal extractor $\mathcal{E}^*(\cdot) = \big(\mathcal{E}_1^*(\cdot), \cdots, \mathcal{E}_m^*(\cdot)\big)$, an ideal combiner $\mathcal{C}^*(\cdot)$, and a compositional residual operator $\mathcal{R}(\cdot)$ such that*

$$\mathcal{F}(\boldsymbol{u}) = \mathcal{C}^*\big(\mathcal{E}^*(\boldsymbol{u})\big) + \mathcal{R}(\boldsymbol{u}), \quad \forall \boldsymbol{u} \in \mathcal{U}, \tag{6}$$

*where $\mathcal{U}$ denotes the admissible state set. The residual operator $\mathcal{R}$ represents the part of the target dynamics that is not captured by the selected extractor-combiner representation. It may include hidden variables, boundary-induced terms or other contributions that are not expressed by the chosen intermediate physical components. We assume that this residual is uniformly bounded on $\mathcal{U}$*

$$\sup_{\boldsymbol{u} \in \mathcal{U}} \|\mathcal{R}(\boldsymbol{u})\|_2 \leq \delta_{\text{comp}}. \tag{7}$$

The quantity $\delta_{\text{comp}}$ measures the structural mismatch between the true evolution operator and the selected hierarchical compositional representation. If the selected hierarchy exactly represents the target dynamics, then $\delta_{\text{comp}} = 0$.

Accordingly, our HPE-AFNO model can be represented as:

$$\hat{\mathcal{F}}(\boldsymbol{u}, \theta) = \hat{\mathcal{C}}\big(\hat{\mathcal{E}}(\boldsymbol{u}, \theta_{\hat{\mathcal{E}}}), \theta_{\hat{\mathcal{C}}}\big), \tag{8}$$

where $\hat{\mathcal{E}}(\cdot) = \Big(\hat{\mathcal{E}}_1(\cdot), \cdots, \hat{\mathcal{E}}_m(\cdot)\Big)$ denotes the approximation of $\mathcal{E}^*(\cdot)$ parameterized by $\theta_{\hat{\mathcal{E}}}$, $\hat{\mathcal{C}}(\cdot)$ denotes the approximation of $\mathcal{C}^*(\cdot)$ parameterized by $\theta_{\hat{\mathcal{C}}}$, and $\theta = (\theta_{\hat{\mathcal{E}}}, \theta_{\hat{\mathcal{C}}})$. We further assume that the second-level combiner is Lipschitz continuous with respect to its feature input:

**Assumption 2.** *Let $\mathcal{Z}$ be a subset of the feature space that satisfies $\Big(\mathcal{E}^*(\mathcal{U}) \cup \hat{\mathcal{E}}(\mathcal{U})\Big) \subseteq \mathcal{Z}$. The ideal combiner $\mathcal{C}^*$ is assumed to be Lipschitz continuous on $\mathcal{Z}$. That is, there exists a constant $L_{\mathcal{C}}$ such that, for any $\boldsymbol{z}_1$, $\boldsymbol{z}_2 \in \mathcal{Z}$,*

$$\|\mathcal{C}^*(\boldsymbol{z}_1) - \mathcal{C}^*(\boldsymbol{z}_2)\|_2 \leq L_{\mathcal{C}} \|\boldsymbol{z}_1 - \boldsymbol{z}_2\|_2. \tag{9}$$

To quantify the two-level approximation implied by this factorization, we define three complementary error measures on the state variable region $\mathcal{U}$. First, for each extracted feature channel, we measure its worst-case deviation from the corresponding ideal term by:

$$\epsilon_{\mathcal{E}_i} = \sup_{\boldsymbol{u} \in \mathcal{U}} \left\| \hat{\mathcal{E}}_i(\boldsymbol{u}, \theta_{\hat{\mathcal{E}}}) - \mathcal{E}_i^*(\boldsymbol{u}) \right\|_2, \quad i = 1, \cdots, m, \tag{10}$$

which summarizes the first-level extraction error across $\mathcal{U}$. Second, we quantify the error of the second-level model in learning the governing combination rule:

$$\epsilon_{\mathcal{C}} = \sup_{\boldsymbol{z} \in Z} \left\| \hat{\mathcal{C}}(\boldsymbol{z}, \theta_{\hat{\mathcal{C}}}) - \mathcal{C}^*(\boldsymbol{z}) \right\|_2. \tag{11}$$

Finally, to connect the hierarchical analysis to the learning objective, we define the overall approximation error as:

$$\epsilon = \sup_{\boldsymbol{u} \in \mathcal{U}} \left\| \hat{\mathcal{F}}(\boldsymbol{u}, \theta) - \mathcal{F}(\boldsymbol{u}) \right\|_2. \tag{12}$$

Together, $\{\epsilon_{\mathcal{E}_i}\}_{i=1}^m$, $\epsilon_{\mathcal{C}}$, and $\epsilon$ provide an interpretable accounting of model error: $\{\epsilon_{\mathcal{E}_i}\}_{i=1}^m$ characterizes the error when extracting fundamental physical terms, $\epsilon_{\mathcal{C}}$ captures deviations in their governing composition, and $\epsilon$ measures the resulting overall discrepancy in the induced dynamics. Then we have the following theorem of error decomposition.

**Theorem 1.** *Under Assumptions 1 and 2, the overall error admits the following decomposition-bound:*

$$\epsilon \leq \epsilon_{\mathcal{C}} + L_{\mathcal{C}} \left( \sum_{i=1}^{m} \epsilon_{\mathcal{E}_i}^2 \right)^{\frac{1}{2}} + \delta_{\text{comp}}. \tag{13}$$

The proof is provided in Supplementary Note 6. Theorem 1 shows that the total approximation error is controlled by three quantities: the second-level combination error, the accumulated errors in the extracted first-level physical components scaled by the Lipschitz constant of the ideal combiner, and the compositional residual that measures the mismatch between the selected hierarchy and the true evolution operator. This decomposition clarifies the mechanism by which hierarchy can improve error control when the assumed compositional structure is informative. A single-level model is required to approximate the high-order nonlinear operator within one mapping, so errors associated with constitutive components, differential operators, nonlinear couplings and unresolved effects are entangled. In contrast, the hierarchical model separates these sources into first-level extraction, second-level combination and residual structural mismatch. When the decomposition is well aligned with the organization of the underlying PDE, the approximation burden is redistributed into simpler subproblems, while the remaining discrepancy is explicitly accounted for by $\delta_{\text{comp}}$.

This perspective also clarifies the role of physics embedding. If a subset of the physical components is known and directly injected into the first level, the corresponding feature errors $\epsilon_{\mathcal{E}_i}$ are removed or substantially reduced from the learned part of the model. If part of the governing combination rule is known and imposed in the second level, the combination error $\epsilon_{\mathcal{C}}$ is reduced to the unresolved interactions that still need to be learned. Thus, in partially specified systems, the advantage of HPE-AFNO is not merely that it can approximate the target operator, but that it can factor out available physical knowledge and leave a smaller unresolved component to be learned.

### 5.3 Parameter-complexity advantage of hierarchical Fourier approximation

The hierarchical error decomposition above shows that the approximation error of HPE-AFNO can be separated into first-level physical-expression extraction error, second-level governing-combination error and a compositional residual. We now use this decomposition to address a complementary question: how many learnable Fourier parameters are required to achieve a prescribed approximation accuracy? This analysis provides a parameter-complexity perspective on the advantage of a hierarchical model over a single-level operator model. A single-level AFNO needs to approximate the full evolution operator directly, whereas HPE-AFNO decomposes the task into physical-expression extractors and a governing combiner, each of which admits a simpler Fourier approximation when the hierarchy is aligned with the compositional structure of the dynamics. To make the comparison analytically transparent, we focus on the core Fourier mechanism shared by FNO-type architectures, namely truncated frequency-domain mixing, rather than on implementation-specific details of the full AFNO block.

Throughout this section, we assume that the spatial domain is the $d$-dimensional periodic torus $\Omega = T^d \triangleq [0, 2\pi)^d$ so that all spatial fields are treated as periodic functions on $\Omega$. Let $e_{\boldsymbol{\xi}}(\boldsymbol{x}) = e^{i\boldsymbol{\xi}\cdot\boldsymbol{x}}$, $\boldsymbol{\xi} \in \mathbb{Z}^{\boldsymbol{d}}$, denote the Fourier basis on $\Omega$. For any scalar field $u \in L^2(\Omega)$, we write its Fourier expansion as

$$u(\boldsymbol{x}) = \sum_{\boldsymbol{\xi}\in\mathbb{Z}^{\boldsymbol{d}}} \hat{u}(\boldsymbol{\xi}) e_{\boldsymbol{\xi}}(\boldsymbol{x}), \tag{14}$$

where

$$\hat{u}(\boldsymbol{\xi}) = \frac{1}{(2\pi)^d} \int_{\Omega} u(\boldsymbol{x}) e^{-i\boldsymbol{\xi}\cdot\boldsymbol{x}} d\boldsymbol{x}. \tag{15}$$

We consider a translation-invariant linear operator $\mathcal{L}$. Since $\mathcal{L}$ is translation invariant, it is diagonalized by the Fourier basis and admits a Fourier symbol $\sigma_{\mathcal{L}}: \mathbb{Z}^{\boldsymbol{d}} \to \mathbb{C}$, such that

$$\mathcal{L}(u)(\boldsymbol{x}) = \sum_{\boldsymbol{\xi} \in \mathbb{Z}^{\boldsymbol{d}}} \sigma_{\mathcal{L}}(\boldsymbol{\xi}) \hat{u}(\boldsymbol{\xi}) e_{\boldsymbol{\xi}}(\boldsymbol{x}). \tag{16}$$

For example, for the differential operator $\partial^{\alpha}$ with $d = 1$, the corresponding Fourier symbol is $\sigma_{\partial^{\alpha}}(\xi) = (i\xi)^{\alpha}$. In practice, Fourier operator layers retain only a finite number of frequency modes. Accordingly, a truncated Fourier mixing layer with truncation level $K$ acts as:

$$\mathcal{L}_K^W(u)(\boldsymbol{x}) = \sum_{\|\boldsymbol{\xi}\|_\infty \le K} W(\boldsymbol{\xi}) \hat{u}(\boldsymbol{\xi}) e_{\boldsymbol{\xi}}(\boldsymbol{x}), \tag{17}$$

where $W: \mathbb{Z}^{\boldsymbol{d}} \to \mathbb{C}$ denotes the learnable frequency-domain weight, set to zero outside the truncated frequency cube $\{\boldsymbol{\xi} \in \mathbb{Z}^{\boldsymbol{d}}: \|\boldsymbol{\xi}\|_\infty \le K\}$. For approximating $\mathcal{L}$, the optimal truncated weight is $W^*(\boldsymbol{\xi}) = \sigma_{\mathcal{L}}(\boldsymbol{\xi}) \mathbf{1}_{\{\|\boldsymbol{\xi}\|_\infty \le K\}}$ and the corresponding optimal truncated Fourier approximation is denoted by $\mathcal{L}_K^* = \mathcal{L}_K^{W^*}$. The number of retained Fourier modes is $N_K = (2K+1)^d$, which determines the learnable parameter scale of the frequency-domain mixing component up to fixed channel and block-structure constants.

To quantify the truncation error of this Fourier mixing component, we impose the following spectral regularity assumption on the target operator and the input class.

**Assumption 3.** *The translation-invariant operator $\mathcal{L}$ and the input class $\mathcal{U}$ satisfy the following two conditions. First, the Fourier symbol of $\mathcal{L}$ has at most polynomial growth. That is, there exist constants $C_{\mathcal{L}} > 0$ and $p \ge 0$ such that*

$$|\sigma_{\mathcal{L}}(\boldsymbol{\xi})| \le C_{\mathcal{L}}(1 + \|\boldsymbol{\xi}\|_\infty)^p, \quad \boldsymbol{\xi} \in \mathbb{Z}^{\boldsymbol{d}}. \tag{18}$$

*Second, the input fields have controlled high-frequency decay. That is, there exist constants $M_{\mathcal{U}} > 0$ and $r > 0$ such that*

$$\sup_{u \in \mathcal{U}} \sum_{\boldsymbol{\xi} \in \mathbb{Z}^{\boldsymbol{d}}} (1 + \|\boldsymbol{\xi}\|_\infty)^{2(p+r)} |\hat{u}(\boldsymbol{\xi})|^2 \le M_{\mathcal{U}}^2. \tag{19}$$

Here $p$ specifies an upper bound on the polynomial growth of the Fourier symbol, while $r$ measures the additional spectral decay of the input class beyond that required to compensate for this growth. Under this assumption, a larger admissible $r$ yields a faster truncation rate for the Fourier approximation. These conditions allow the Fourier truncation error to be bounded in terms of the retained frequency level $K$, thereby yielding an explicit parameter-scaling law for truncated Fourier mixing.

**Theorem 2.** *Under Assumption 3, the optimal truncated Fourier approximation $\mathcal{L}_K^*$ satisfies*

$$\sup_{u\in\mathcal{U}}\|\mathcal{L}(u)-\mathcal{L}_K^*(u)\| \le C_{\mathcal{L}}M_{\mathcal{U}}(1+K)^{-r}, \tag{20}$$

*where* $\|\mathcal{L}(u)-\mathcal{L}_K^*(u)\| \triangleq \left(\frac{1}{(2\pi)^d}\int_\Omega |\mathcal{L}(u)(\boldsymbol{x})-\mathcal{L}_K^*(u)(\boldsymbol{x})|^2 d\boldsymbol{x}\right)^{\frac{1}{2}}$*.Consequently, for any prescribed component error tolerance* $\eta>0$*, it is sufficient to choose a truncation level*

$$K=\left\lceil\left(\frac{C_{\mathcal{L}}M_{\mathcal{U}}}{\eta}\right)^{\frac{1}{r}}-1\right\rceil. \tag{21}$$

*to ensure* $\|\mathcal{L}(u)-\mathcal{L}_K^*(u)\|\le\eta$*.*

The proof is provided in Supplementary Note 6. Since the number of retained Fourier modes is $N_K=(2K+1)^d$, the number of learnable frequency-domain parameters required to make the approximation error no larger than $\eta$ satisfies $N_K(\eta)\le\alpha_{\mathcal{L}}\eta^{-d/r}$ where $\alpha_{\mathcal{L}}>0$ is independent of $\eta$ and absorbs the dimension-dependent mode-counting constant as well as fixed channel and block-structure factors.

Theorem 2 quantifies a sufficient parameter-scaling law for truncated Fourier mixing. It shows that the effective approximation rate $r$ controls how rapidly the sufficient number of Fourier parameters grows as the target error tolerance becomes smaller. Although Theorem 2 is stated for translation-invariant linear Fourier multipliers, it provides a prototype for the Fourier approximation rates used in the hierarchical complexity analysis below. We therefore introduce the following effective approximation-rate assumption for the first-level extractors, the second-level combiner and the single-level baseline operator.

**Assumption 4.** *For the i-th first-level physical-expression extractor $\mathcal{E}_i^*$, there exist constants $r_{\mathcal{E}_i}>0$ and $\alpha_{\mathcal{E}_i}>0$, independent of the target error $\eta_{\mathcal{E}_i}$, such that approximating $\mathcal{E}_i^*$ with error tolerance $\eta_{\mathcal{E}_i}$ in the $L^2$-based error requires no more than*

$$N_{\mathcal{E}_i}(\eta_{\mathcal{E}_i})\le\alpha_{\mathcal{E}_i}\eta_{\mathcal{E}_i}^{-\frac{d}{r_{\mathcal{E}_i}}}, \tag{22}$$

*learnable Fourier parameters. Similarly, for the second-level governing combiner $\mathcal{C}^*$, there exist constants $r_{\mathcal{C}}>0$ and $\alpha_{\mathcal{C}}>0$, independent of $\eta_{\mathcal{C}}$, such that approximating $\mathcal{C}^*$ with error tolerance $\eta_{\mathcal{C}}$ requires no more than*

$$N_{\mathcal{C}}(\eta_{\mathcal{C}}) \leq \alpha_{\mathcal{C}} \eta_{\mathcal{C}}^{-\frac{d}{r_{\mathcal{C}}}}. \quad (23)$$

*For comparison, the single-level AFNO baseline that directly approximates the full evolution operator is assumed to admit an effective Fourier approximation rate* $r_{\mathcal{S}} > 0$*. That is, there exists a constant* $\alpha_{\mathcal{S}} > 0$*, independent of* $\epsilon$*, such that achieving overall error tolerance* $\epsilon$ *requires no more than*

$$N_{SL}(\epsilon) \leq \alpha_{\mathcal{S}} \epsilon^{-\frac{d}{r_{\mathcal{S}}}}. \quad (24)$$

*learnable Fourier parameters. Here, all errors* $\eta_{\mathcal{E}_i}$*,* $\eta_{\mathcal{C}}$ *and* $\epsilon$ *are measured in the same* $L^2$*-based error sense as the approximation errors used in Theorem 1. The constants* $\alpha_{\mathcal{E}_i}$*,* $\alpha_{\mathcal{C}}$ *and* $\alpha_{\mathcal{S}}$ *collect the corresponding mode-counting, channel and block-structure factors, whereas* $r_{\mathcal{E}_i}$ *,* $r_{\mathcal{C}}$ *and* $r_{\mathcal{S}}$ *characterize the effective Fourier approximation rates of the extractor, combiner and single-level evolution operators, respectively.*

Combining the effective Fourier approximation rates in Assumption 4 with the hierarchical error decomposition in Theorem 1, we obtain a sufficient Fourier-parameter budget for HPE-AFNO to achieve a prescribed overall error tolerance and compare it with the corresponding single-level AFNO scaling.

**Theorem 3.** *Under Assumption 4, for any target error tolerance* $\epsilon > 2\delta_{\text{comp}}$*, there exists an HPE-AFNO approximation whose total error is no larger than* $\epsilon$*, with sufficient learnable Fourier parameter count bounded by*

$$N_{\text{HPE}}(\epsilon) \leq \sum_{i=1}^{m} \alpha_{\mathcal{E}_i} \left( \frac{4L_{\mathcal{C}}\sqrt{m}}{\epsilon} \right)^{\frac{d}{r_{\mathcal{E}_i}}} + \alpha_{\mathcal{C}} \left( \frac{4}{\epsilon} \right)^{\frac{d}{r_{\mathcal{C}}}}. \quad (25)$$

The proof is provided in Supplementary Note 6. In the simplified case where $r_{\mathcal{E}} = \min_{1 \leq i \leq m} r_{\mathcal{E}_i}$ and $\alpha_{\mathcal{E}} = \max_{1 \leq i \leq m} \alpha_{\mathcal{E}_i}$, we have

$$N_{\text{HPE}}(\epsilon) \leq m\alpha_{\mathcal{E}} \left( \frac{4L_{\mathcal{C}}\sqrt{m}}{\epsilon} \right)^{\frac{d}{r_{\mathcal{E}}}} + \alpha_{\mathcal{C}} \left( \frac{4}{\epsilon} \right)^{\frac{d}{r_{\mathcal{C}}}}. \quad (26)$$

When the selected hierarchy is aligned with the compositional structure of the governing dynamics, the extractor and combiner subproblems admit higher effective Fourier approximation rates than the full single-level operator. In the regime $\min\{r_{\mathcal{E}}, r_{\mathcal{C}}\} > r_{\mathcal{S}}$, the sufficient parameter complexity of HPE-AFNO grows more slowly than that of the single-level model as the target error tolerance becomes smaller. In particular, when $\delta_{\mathrm{comp}}$ is sufficiently small relative to the target error (i.e., $\delta_{\mathrm{comp}} = o(\epsilon)$)

$$\frac{N_{\mathrm{HPE}}(\epsilon)}{N_{\mathrm{SL}}(\epsilon)} \to 0, \quad \epsilon \to 0. \tag{27}$$

This result establishes the parameter-complexity advantage of hierarchical Fourier approximation. For a prescribed target error, a single-level AFNO allocates Fourier parameters to approximate the full evolution operator as one entangled mapping. In contrast, HPE-AFNO distributes the approximation task across first-level physical-expression extractors and a second-level governing combiner. When the hierarchy is aligned with the compositional structure of the dynamics, these component operators admit higher effective Fourier approximation rates than the full single-level operator. Therefore, under the same approximation error, the sufficient number of learnable Fourier parameters required by HPE-AFNO can be lower than that required by a single-level AFNO. This parameter-complexity reduction provides the theoretical basis for using hierarchical physics embedding as a more efficient Fourier operator approximation strategy.

### 5.4 Equation discovery with partial structural knowledge

The proposed methodology for identifying unknown physical expressions under partially specified PDE structures comprises three steps: (1) spatiotemporal dynamics prediction, (2) concentration binning analysis, and (3) deep symbolic regression. These steps are detailed below.

**Spatiotemporal dynamics prediction.** A key challenge in equation discovery with partial structural knowledge is that target physical expressions, such as concentration-dependent diffusivity and homogeneous chemical potential in phase-field systems, are typically not directly measurable. Furthermore, experimentally accessible state variables, such as concentrations ($c$), are typically sparse and corrupted by noise. Therefore, a crucial initial step involves processing this raw, sparse data to reconstruct a high-fidelity, denoised representation of the system's dynamics, effectively inferring the behavior of the underlying, unmeasured physical terms.

Within our framework, the HPE-AFNO serves as this reconstruction model. This procedure adheres to the approach outlined previously in the 'HPE-AFNO framework' section. Specifically, we configure the HPE-AFNO model by embedding known fundamental physical operators (e.g., diffusion terms) into its first hierarchical level and known composite terms (e.g., combinations of derivatives) into the second level. This structured model is then trained using the available sparse measurements of the state variable $u$ (i.e., concentration in this work).

Furthermore, a critical physical requirement is that the inferred unknown terms exhibit a continuous functional dependence on the input state variable (concentration). To enforce this, we integrate a kernel-based feature mapping mechanism into the HPE-AFNO model tasked with learning these unknown physical terms. The underlying principle is that unknown physical terms or constitutive relations (e.g., diffusivity $D(c)$ or chemical potential $\mu(c)$) should yield similar values for similar input concentrations. This is achieved using a positive definite kernel function that quantifies similarity between input values. While various kernels are applicable, we utilize a Gaussian kernel[36] for its effectiveness and simplicity:

$$\kappa\big(u(i), u(j)\big) = \exp\left(-\frac{\big(u(i) - u(j)\big)^2}{2\sigma^2}\right) \tag{28}$$

where $u(i)$ and $u(j)$ are input variable values (e.g., concentrations) at two different spatial locations, and $\sigma$ is a hyperparameter controlling the characteristic length scale of similarity. The kernel $\kappa$ assigns a similarity score approaching 1 for nearby input values, decaying towards 0 as the difference increases.

Normalized consistency weights $w_{\text{cons}}$ are computed as $w_{\text{cons}} = \kappa / \sum(\kappa + \epsilon)$, where the summation is performed over all spatial locations and $\epsilon$ is a small constant ensuring numerical stability. The final encoded features representing the unknown physical term at each location are then calculated as a weighted average of the neural network's outputs across all spatial locations, using these similarity-based weights ($w_{\text{cons}}$). This kernel-based mapping encourages that the predicted unknown terms exhibit continuous dependence on the input variable; locations with similar input values inherently receive similar weighting patterns from the kernel matrix, resulting in similar final encoded features.

**Concentration binning analysis.** Once the spatiotemporal dynamics are reconstructed by the HPE-AFNO, yielding reliable estimates of the unknown physical terms across space and time, the next step is to extract their functional dependence on the state variable (e.g., concentration $c$). Using the CH equation as an illustrative example, we aim to determine the explicit functional forms of terms like the concentration-dependent diffusivity $D(c)$ and the homogeneous chemical potential $\mu_{\text{hom}}(c)$. To achieve this, we employ a concentration binning analysis to the model's inferred values of these unknown terms. This process generates statistical representations that serve as input for the subsequent step of deep symbolic regression.

Specifically, the concentration range (typically normalized to [0, 1]) is divided into $N$ uniform bins. For each bin $i$, corresponding to the concentration interval $[c_i, c_{i+1}]$, we collect all the model's predictions for a given unknown physical term, denoted $\zeta(c_x)$, where the local concentration $c_x$ at spatial location $x$ falls within this interval. The average of these collected predictions $\zeta_i$ is then computed and associated with the midpoint concentration of the bin $(c_i + c_{i+1})/2$. Compiling these average values across all $N$ bins yields a discretized representation of the unknown physical term as a function of concentration (e.g., Fig. 5).

This ensemble averaging within each concentration bin effectively smooths out stochasticity or fluctuations inherent in the neural network's point-wise predictions. It robustly reveals the underlying functional relationship between the unknown physical term and the concentration variable, providing high-quality data suitable for discovering interpretable mathematical expressions via deep symbolic regression.

**Deep symbolic regression.** Having obtained the statistical representations of the unknown physical terms via concentration binning analysis, we can identify potential functional relationships between these terms and the state variables (e.g., the concentration field in the CH equation). This facilitates the application of deep symbolic regression (DSR) to discover the explicit analytical structure of the underlying PDE expressions (e.g., the explicit equations for the diffusivity coefficient $D(c)$ and homogeneous chemical potential $\mu_{\text{hom}}(c)$).

DSR[35] is a widely used technique for data-driven equation discovery that represents mathematical expressions as symbolic trees, which are then typically traversed sequentially. Specifically, DSR often utilizes a recurrent neural network (RNN)[37] to probabilistically generate these expression trees token by token. Let $\tau$ denote the candidate expression. The $i$th token in the traversal is $\tau_i$, selected from a predefined library $\mathcal{L}$ of mathematical variables and operators (e.g., $\{+, -, \times, \div, \log, \exp\}$). The total length of the traversal is $|\tau| = T$.

The RNN, parameterized by $\phi$, outputs a probability distribution $\psi^{(i)}$ at step $i$ for selecting the $i$th token $\tau_i$, conditioned on the preceding tokens $\tau_{1:(i-1)}$. This generative process continues until a complete expression tree is formed. The overall likelihood of sampling expression $\tau$ is given by the product of the probabilities of selecting each token:

$$p(\tau|\phi) = \prod_{i=1}^{T} p(\tau_i | \tau_{1:(i-1)}; \phi) = \prod_{i=1}^{T} \psi_{\mathcal{L}(\tau_i)}^{(i)} \tag{29}$$

Due to the non-differentiability of the reward function with respect to the RNN parameters $\phi$, reinforcement learning (RL)[38] is typically employed for training. In this RL framework, the distribution over expressions $p(\tau|\phi)$ represents the policy, the sequence of sampled tokens forms an action sequence (or episode), and a reward function evaluates the quality of the generated expression $\tau$. Notably, the total reward $R(\tau)$ is typically assigned based on the evaluation of the completed expression (e.g., its fit to the target data), rather than as a sum of time-discounted rewards for individual token selections.

To enhance the discovery of high-performing expressions, a risk-seeking policy gradient objective is adopted:

$$J_{risk}(\phi, \epsilon) = \mathbb{E}_{\tau \sim p(\tau|\phi)}[R(\tau)|R(\tau) \geq q_\epsilon(R)] \quad (30)$$

where the objective, characterized by the parameter $\epsilon$ ($0 < \epsilon \leq 1$), prioritizes maximizing rewards within the top $\epsilon$ quantile ($q_\epsilon(R)$) of the reward distribution.

As illustrated in Fig. 5, we establish potential functional relationships between the inferred unknown physical terms and the concentration field. Accordingly, the DSR reward function $R(\tau)$ quantifies the agreement between a candidate expression τ and the statistical data derived from the concentration binning analysis. This is typically achieved using a metric of normalized root-mean-square error (NRMSE): $R(\tau) = 1/(1 + \text{NRMSE})$. Compared to traditional equation discovery approaches reliant on sparse regression[39,40], symbolic-regression methods can explore mathematical expressions without prescribing a fixed candidate library[41]. The DSR-based method adopted here therefore offers greater flexibility, as it does not require a *priori* assumptions about the underlying equation's structure (further discussion in Supplementary Note 5).

**Acknowledgements**

This work was supported by the National Key Research and Development Program of China (2025YFE0207400), the Tsinghua-Toyota Joint Research Fund, the National Natural Science Foundation of China (62273197), the Beijing Natural Science Foundation (L233027), and the 111 International Collaboration Project (B25027).

**Author Contributions**

X.W. and B.J. conceived the research. X.W. and X.S. performed the modeling and numerical experiments. X.W., X.S., Q.J., Q. Z., H.Z., H.S., and B.J. interpreted the results. All authors edited and reviewed the manuscript. B.J. supervised the work.

**Competing Interests Statement**

The authors declare no competing interests.

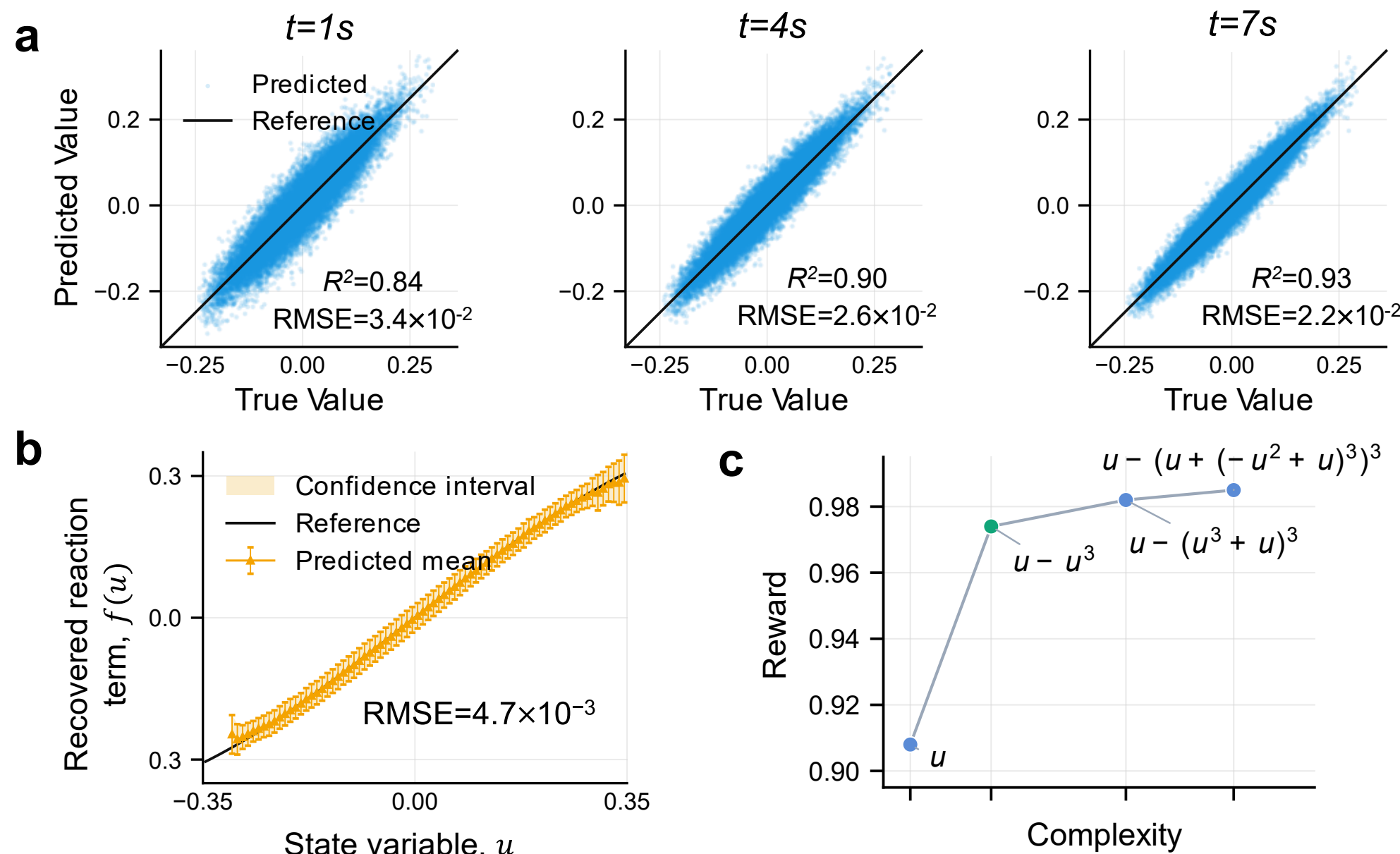

**Extended Data Fig. 1 | Validation of nonlinear reaction discovery in three-dimensional Allen–Cahn dynamics. a**, Predicted-versus-reference comparison of reconstructed three-dimensional Allen–Cahn fields at representative time points, with $R^2$ and RMSE reported for each scatter plot. **b**, Recovered nonlinear reaction term $f(u)$ as a function of the state variable $u$; the predicted mean and confidence interval are compared against the reference reaction $u - u^3$. **c**, Reward–complexity curve from symbolic regression, where complexity is the weighted count of variables, constants and operators in the expression tree[35]. The expression $u - u^3$ is identified as a compact, physically interpretable, high-reward candidate.

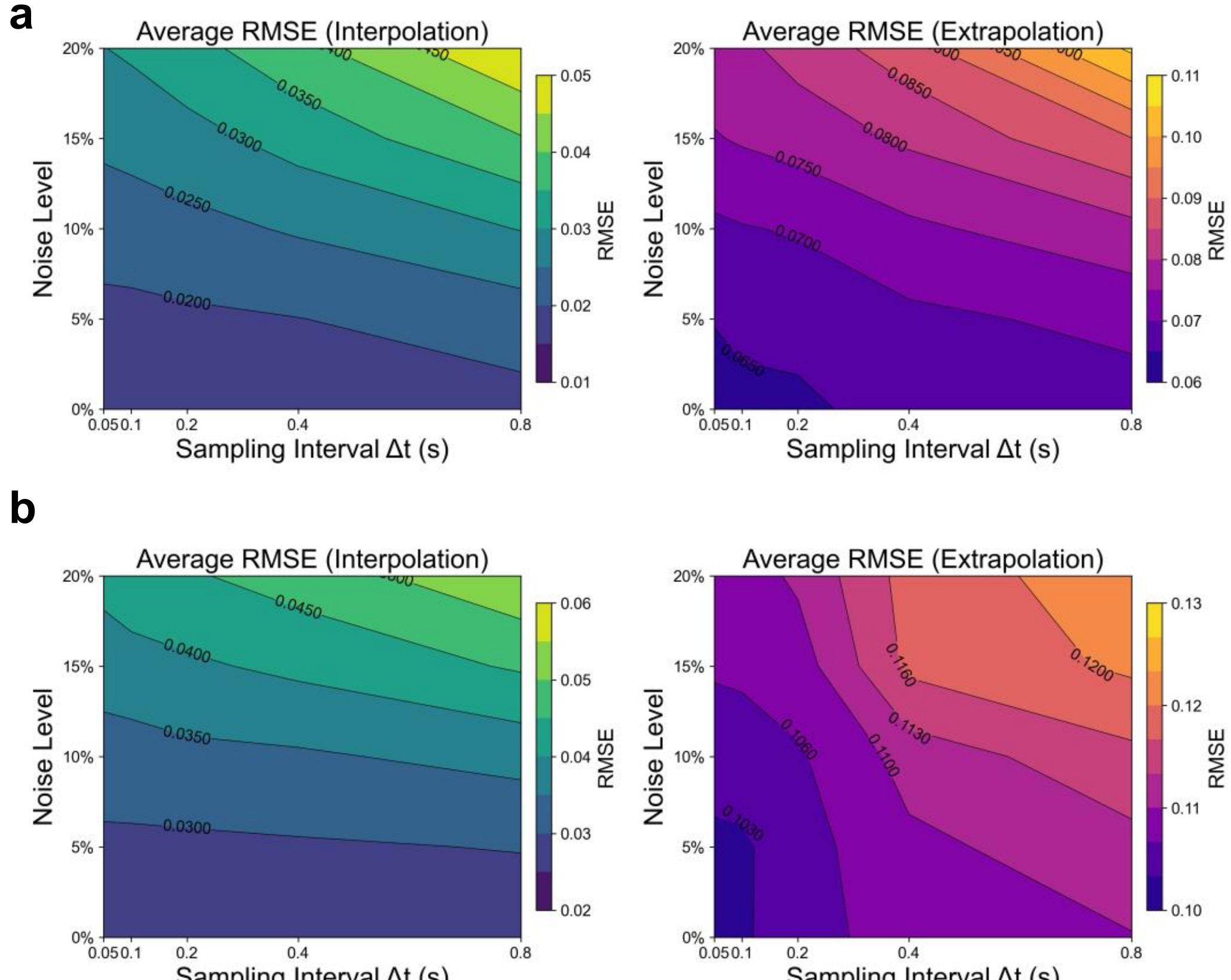

**Extended Data Fig. 2 | Robustness of HPE-AFNO to data sparsity and noise in phase-separation PDE systems. a**, Cahn–Hilliard system; **b**, Allen–Cahn system. Each panel shows contours of the average RMSE for the interpolation (left) and extrapolation (right) regimes across a two-dimensional space spanned by the temporal sampling interval ($\delta t$) and noise level, averaged over five runs with distinct random initial conditions. In both systems, prediction error rises only modestly as sparsity and noise increase, across both regimes.

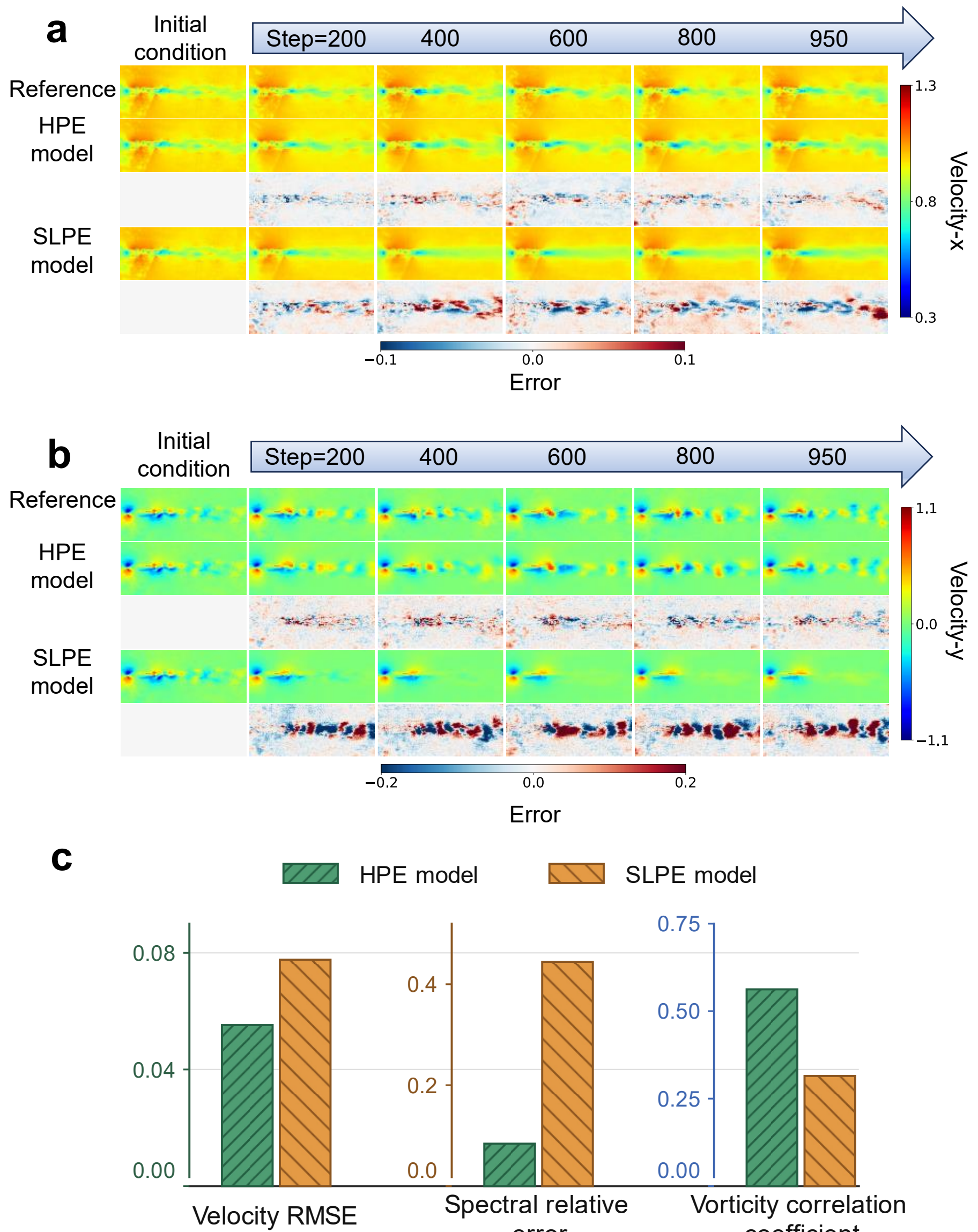

**Extended Data Fig. 3 | Hierarchical physics embedding improves the use of physical priors in wake prediction. a, b,** Temporal rollout of the streamwise velocity $u$ and transverse velocity $v$ from step 200 to step 950, comparing the reference against the single-level physics-embedding (SLPE) and hierarchical physics-embedding (HPE) models, with corresponding error maps. Color bars report field values and signed errors on the evaluation-data scale; no physical units are assigned. In SLPE, physics-derived quantities are concatenated with the raw velocity fields as additional input channels; in HPE, the same quantities enter through hierarchical physical pathways before being integrated with learned spatiotemporal features. For both components, HPE better preserves downstream wake morphology and coherent vortex structures, whereas SLPE accumulates larger errors and distorts small-scale patterns more severely. **c,** Quantitative comparison using velocity RMSE, spectral relative error and vorticity correlation coefficient: lower RMSE and spectral error indicate more accurate fields and better spectral consistency, while a higher vorticity correlation indicates better-preserved rotational dynamics. The lower errors and higher vorticity correlation of HPE show that hierarchical organization provides a more effective inductive bias than input-level concatenation.

**Supplementary Information**

# Hierarchical Physics-Embedded Learning for Partially Known Spatiotemporal Dynamics

Xizhe Wang[1,2,3*], Xiaobin Song[1,2,3*], Hongbo Zhao[4], Qingshan Jia[1,2,3], Qianchuan Zhao[1,2,3], Hao Sun[5], and Benben Jiang[1,2,3†]

1 Department of Automation, BNRist, Tsinghua University, Beijing 100084, China

2 Beijing Key Laboratory of Embodied Intelligence Systems, Beijing 100084, China

3 Institute for Embodied Intelligence and Robotics, Tsinghua University, Beijing 100084, China

4 Department of Physics, Department of Chemistry and Biochemistry, University of California, San Diego, CA 92093, USA

5 Gaoling School of Artificial Intelligence, Renmin University of China, Beijing 100086, China

* These authors contributed equally to this work

† Corresponding authors

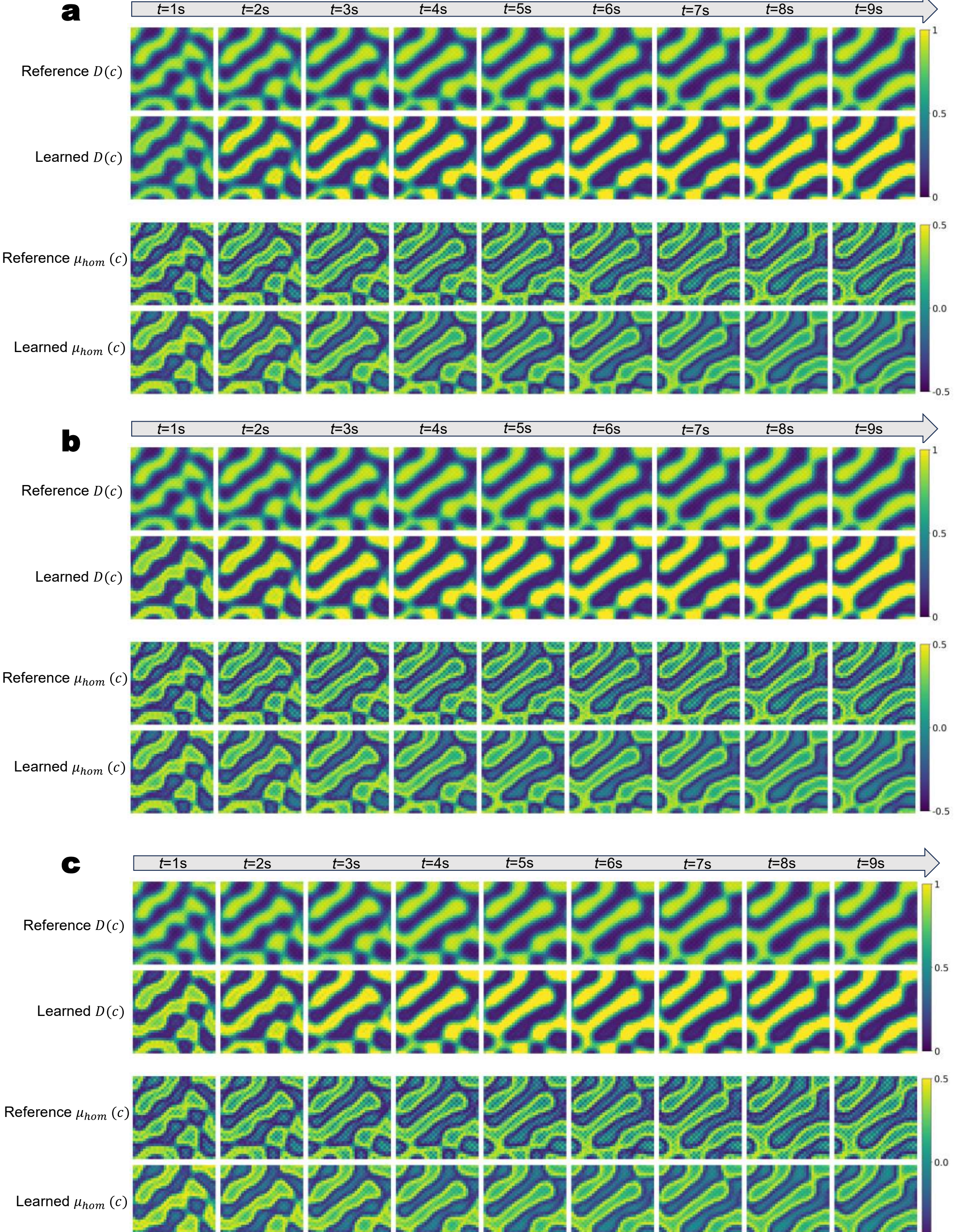

**Supplementary Fig. 1 | Predicted evolution of unknown physical terms under varying noise levels.** Concentration-dependent diffusivity $D(c)$ and homogeneous chemical potential $\mu_{hom}(c)$ predicted by the HPE-AFNO model under three levels of additive Gaussian noise in the measurement data: (a) 0%, (b) 5%, and (c) 10%. For each case, the top row displays the ground truth, while the bottom row shows the corresponding predictions from the first level of the model, where two separate AFNO channels are respectively assigned to learn $D(c)$ and $\mu_{hom}(c)$. Temporal evolution is shown from left to right, spanning $t = 1$s to $t = 9$s. The model maintains high predictive accuracy across noise levels, enabling reliable learning of the physical terms and supporting subsequent concentration binning analysis and deep symbolic regression for equation discovery.

.

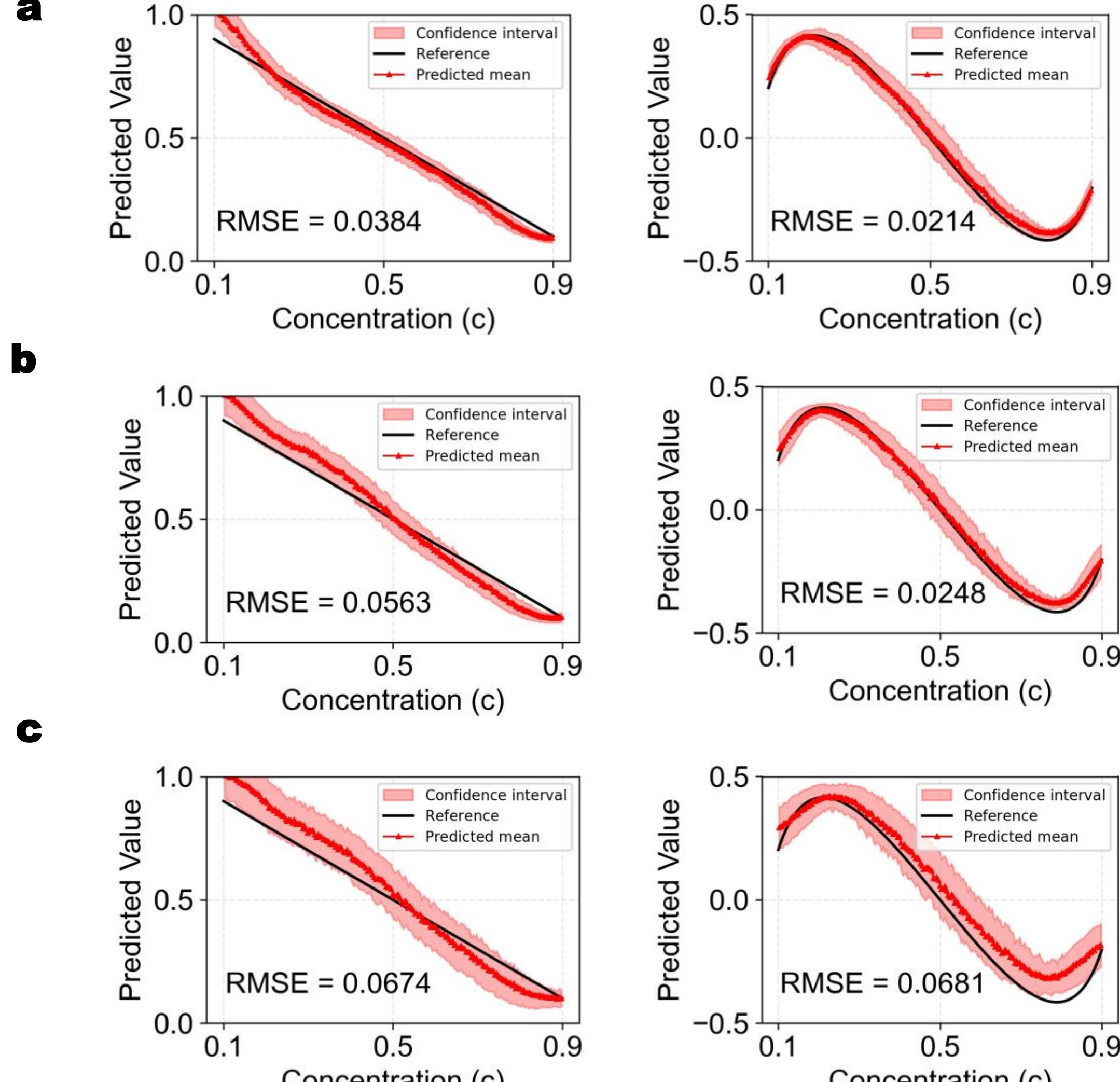

**Supplementary Fig. 2 | Extraction of functional dependence via concentration binning analysis under varying noise levels.** Statistical representations of the concentration-dependent diffusivity $D(c)$ (left panels) and homogeneous chemical potential $\mu_{hom}(c)$ (right panels), derived from the model predictions shown in Supplementary Fig. 1 using concentration binning analysis. Results are presented for three levels of additive Gaussian noise: (a) 0%, (b) 5%, and (c) 10%. For each subplot, model predictions were grouped into discrete concentration intervals and averaged to obtain a smoothed profile of the corresponding physical term, as described in Methods. Black curves represent the ground-truth functions, red curves indicate the predicted means, and shaded bands denote one standard deviation. RMSE values quantify the deviation from the reference expressions. These results demonstrate that the proposed framework consistently extracts accurate functional dependencies across varying noise levels, providing robust inputs for symbolic regression and subsequent closed-form PDE discovery.

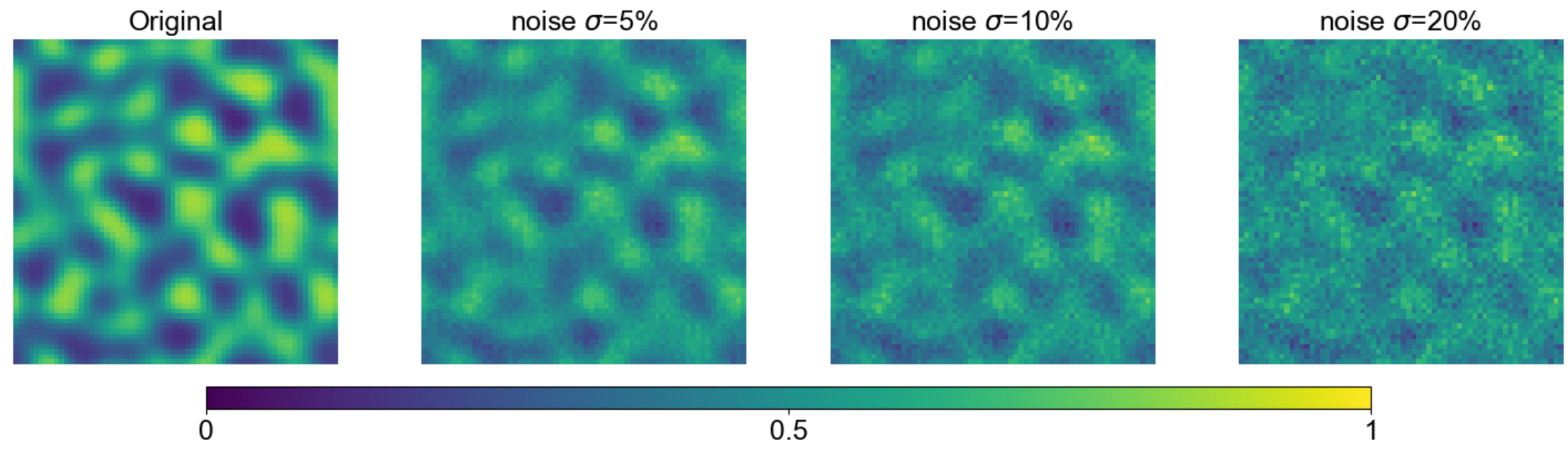

**Supplementary Fig. 3 | Ground truth data with Gaussian noise contamination at varying levels.** The figure illustrates the ground truth data ($\sigma = 0$) and its noisy variants with contamination levels of 5% ($\sigma = 0.05$), 10% ($\sigma = 0.1$), and 20% ($\sigma = 0.2$). The noise intensity increases progressively from left to right.

**Supplementary Note 1: Implementation details of HPE-AFNO**

We propose hierarchical physics-embedded adaptive Fourier neural operator (HPE-AFNO) to investigate the nonlinear dynamical system in which spatiotemporal evolution is governed by a partial differential equation (PDE). Let $\boldsymbol{u}(\boldsymbol{x},t)$ denote the state variable (observation data or snapshots collected from experiments or nature), $\boldsymbol{u}_t$ represent the first-order time derivative of the state variable, and $\boldsymbol{u}_{\boldsymbol{x}}$ donate the spatial derivative. Then the PDE can be formulated as:

$$\boldsymbol{u}_t = \mathcal{F}(\boldsymbol{u}, \boldsymbol{u}_x, \boldsymbol{u}_{xx}, \dots, \boldsymbol{x}, t) \tag{S1}$$

where $\mathcal{F}$ is a nonlinear function involving $\boldsymbol{u}$ and its spatial derivatives of various orders. To simulate the process of time evolution and reduce the number of parameters, we employ a recurrent network architecture with shared parameters to solve the PDE following the forward Euler scheme:

$$\widehat{\boldsymbol{u}}^{k+1} = \widehat{\boldsymbol{u}}^{k} + \hat{\mathcal{F}}(\widehat{\boldsymbol{u}}^{k}, \theta)\delta t \tag{S2}$$

where $\widehat{\boldsymbol{u}}^{k}$ represents the prediction at time $t_k$, and $\hat{\mathcal{F}}$ is an approximation of $\mathcal{F}$ parameterized by $\theta$. Our HPE-AFNO model is designed for $\hat{\mathcal{F}}$.

For the core architecture of HPE-AFNO, we design a hierarchical architecture that separately approximates low-order spatial features and their combinations to reduce the complexity and improve the accuracy of learning spatial evaluation, as shown in **Fig.1c**, which can be represented as:

$$\hat{\mathcal{F}}(\widehat{\boldsymbol{u}}_{\boldsymbol{k}}, \theta) = \hat{\mathcal{C}}\left(\hat{\mathcal{E}}(\widehat{\boldsymbol{u}}^{k}, \theta_{\hat{\mathcal{E}}}), \theta_{\hat{\mathcal{C}}}\right) \tag{S3}$$

where $\hat{\mathcal{E}}(\cdot) = \left(\hat{\mathcal{E}}_1(\cdot), \cdots, \hat{\mathcal{E}}_m(\cdot)\right)$ denotes the approximation of $m$ fundamental physical terms parameterized by $\theta_{\hat{\mathcal{E}}}$, $\hat{\mathcal{C}}(\cdot)$ denotes the approximation of the rule that combines these terms into the time derivative parameterized by $\theta_{\hat{\mathcal{C}}}$, and $\theta = (\theta_{\hat{\mathcal{E}}}, \theta_{\hat{\mathcal{C}}})$.

This hierarchical framework improves the ability to learn higher-order differential operators, aligns with the formulation of physical laws, and facilitates the integration of known physical information. For known physical terms, we can directly embed them into our model, which helps speed up training and enhance prediction accuracy. Specifically, known spatial features can be embedded by introducing an additional channel in the first-level AFNO, while known combinations can be integrated by constructing a separate channel that impacts the features extracted by the first-level AFNO, as illustrated in **Fig. 1c**. Notably, we employ first-order difference method to approximate first differentiation considering the speed of training and prediction. For unknown terms or combination relationships, we choose AFNO [1] to extract intricate spatial patterns in the frequency domain, as shown in **Fig. 1a** (we use the code in

https://github.com/NVIDIA/modulus/blob/1f89d93da8cfb0e093bb9ad83f0eadde6c1cd5c9/modulus/models/afno/afno.py). A Fourier attention layer in AFNO consists of three parts: (1) employ the discrete Fourier transform to mix different tokens: $\boldsymbol{Z}[i,j,:] = \mathrm{DFT}(\overline{\boldsymbol{U}})[i,j,:]$ where $\overline{\boldsymbol{U}} \in \mathbb{R}^{H' \times W' \times d}$ is obtained through patch and position embedding from $\boldsymbol{U} \in \mathbb{R}^{H \times W}$; (2) multiply the weight matrix $\boldsymbol{W}_f \in \mathbb{C}^{H' \times W' \times d \times d}$ in the frequency domain to mix different channels: $\overline{\boldsymbol{Z}}[i,j,:] = \boldsymbol{W}_f[i,j,:,:]\boldsymbol{Z}[i,j,:]$; and (3) adopt the inverse discrete Fourier transform: $\boldsymbol{Y}[i,j] = \mathrm{IDFT}(\overline{\boldsymbol{Z}})[i,j]$. Furthermore, in the implementation of AFNO, the Fourier attention layer introduces some tricks. For the weight matrix $\boldsymbol{W}_f$ in step (2), the layer selects a block-diagonal structure that divides $\boldsymbol{W}_f[i,j,:,:]$ into $k$ weight blocks with the size of $(d/k) \times (d/k)$ to reduce the number of parameters, allow parallel computation, and achieve the effect of multi-head attention mechanism [1]. To further reduce the number of learnable parameters and improve the adaptability, the multilayer perceptron (MLP) with shared parameters is adopted in step (2): $\overline{\boldsymbol{Z}}[i,j,:] = \boldsymbol{W}_{f2}\sigma\left(\boldsymbol{W}_{f1}\boldsymbol{Z}[i,j,:]\right) + \boldsymbol{b}$ [1]. Moreover, AFNO also employs the soft-thresholding and shrinkage operation: $\overline{\boldsymbol{Z}}[i,j,:] = \mathrm{Softshrink}_{\lambda}\left(\boldsymbol{W}_f[i,j,:,:]\boldsymbol{Z}[i,j,:]\right)$, where $\mathrm{Softshrink}_{\lambda}(x) = \mathrm{sign}(x)\max\{|x| - \lambda, 0\}$, to sparsify the tokens and increase attention to the important ones [1]. Here $\lambda$ is a hyperparameter controlling the sparsity. When training HPE-AFNO, we utilize the Mean Square Error (MSE) as loss function and employ the Adam optimizer.

**Supplementary Note 2: Baseline methods**

In this work, we used four baseline models for comparison: Physics-encoded Recurrent Convolutional Neural Network (PeRCNN) [2], Physics-encoded Spectral Attention Network (PeSANet) [3], Fourier Neural Operator (FNO) [4] and Laplace Neural Operator (LNO) [5]. PeRCNN and PeSANet were used as physics-encoded baselines, while FNO and LNO were used as data-driven neural operator baselines. All baseline models were trained using the same training data and evaluated under the same autoregressive rollout setting as HPE-AFNO.

***Physics-encoded Recurrent Convolutional Neural Network (PeRCNN)***. PeRCNN is a recurrent neural architecture designed to learn spatiotemporal dynamics with embedded physical structures [2]. It predicts the temporal evolution of the state variable through a forward-Euler update, where the state increment is represented by a trainable convolutional module. The key component of PeRCNN is the Π-block, which uses several parallel convolutional layers to extract local spatial features from the current state. These features are then multiplied element-wise to represent nonlinear interactions, and a $1 \times 1$ convolution is used to combine the resulting channels into the final state increment. In our phase-field experiments, PeRCNN was supplied with the same available local physical operator as HPE-AFNO, such as the Laplacian operator, so that it served as a local physics-encoded recurrent baseline.

***Physics-encoded Spectral Attention Network (PeSANet).*** PeSANet is a physics-encoded neural network that combines local physical encoding with spectral-domain feature learning [3]. It contains a physics-encoded component for incorporating available differential operators and a spectral attention component for modeling nonlocal spatial interactions. The physics-encoded part helps the model use known local physical information, while the spectral part improves its ability to capture long-range dependencies in PDE-governed dynamics. In our experiments, PeSANet was provided with the same available local physical prior as PeRCNN and HPE-AFNO in the phase-field systems. It was therefore used as a physics-aware baseline that combines local physical priors with spectral feature extraction.

***Fourier Neural Operator (FNO).*** FNO is a neural operator that learns mappings between input and output functions through Fourier-domain parameterization [4]. The input field is first lifted to a higher-dimensional feature space and transformed into the frequency domain. A set of low-frequency Fourier modes is then multiplied by learnable complex-valued weights, while the remaining modes are truncated. The transformed features are mapped back to the physical domain by an inverse Fourier transform and further processed to produce the predicted field. Following the autoregressive setting in PDEBench [6], FNO uses the predicted states from the previous three time steps, $\hat{\boldsymbol{u}}^{t-2}$, $\hat{\boldsymbol{u}}^{t-1}$ and $\hat{\boldsymbol{u}}^{t}$, to predict the next state, $\hat{\boldsymbol{u}}^{t+1} = \mathrm{FNO}(\hat{\boldsymbol{u}}^{t-2}, \hat{\boldsymbol{u}}^{t-1}, \hat{\boldsymbol{u}}^{t})$. In this work, FNO was used as a data-driven operator-learning baseline without explicit physical priors.

***Laplace Neural Operator (LNO).*** LNO is a neural operator that represents the input–output mapping in the Laplace domain [5]. Instead of learning the operator through Fourier modes, LNO parameterizes the dynamics using a pole–residue representation after applying the Laplace transform. This formulation is designed to capture transient and non-periodic dynamics in time-dependent systems. To keep the temporal prediction setting consistent with FNO, LNO also uses the predicted states from the previous three time steps to predict the next state. In this work, LNO was used as a second data-driven neural operator baseline without explicit physical priors.

**Supplementary Note 3: Evaluation metrics**

We evaluated the predictive performance of each model using pointwise errors and system-specific physical diagnostics. The pointwise metric measures local agreement with the reference solution, whereas the physical diagnostics assess whether the predicted rollouts preserve the characteristic morphology, energetics, spectra and vortical structures of the underlying system.

Unless stated otherwise, each time-dependent diagnostic was computed for the predicted and reference rollouts at the same time points. Summary errors were then obtained by averaging the corresponding relative error over the evaluation interval used in the analysis. For the Cahn-Hilliard and three-dimensional Allen-Cahn summary bars, this extrapolation interval was $t > 9\mathrm{s}$.

**Root-mean-square error**

For a scalar field $u$, the root-mean-square error (RMSE) at time $t$ was defined as

$$\mathrm{RMSE}(t) = \left[\frac{1}{N}\sum_{i=1}^{N}\left(u_{\mathrm{pred}}(i,t) - u_{\mathrm{ref}}(i,t)\right)^{2}\right]^{\frac{1}{2}}. \tag{S4}$$

Here, $N$ is the number of evaluated spatial grid points, $i$ indexes grid points, $u_{\mathrm{pred}}$ denotes the predicted field and $u_{\mathrm{ref}}$ denotes the reference field. For vector-valued velocity fields, the same definition was applied over the evaluated components and grid points. When a scalar summary was required, the time-dependent RMSE values were averaged over the corresponding temporal interval.

**Cahn-Hilliard phase-separation morphology**

For the two-dimensional Cahn-Hilliard system, we used structure-factor-based diagnostics to quantify the morphology of phase separation. Let $c(x,t)$ denote the concentration field, where $x = (x,y)$ is the spatial coordinate. Before computing the Fourier spectrum, the spatial mean $\bar{c}(t)$ was subtracted from each snapshot. The structure factor was then computed as

$$S(q,t) = |\mathcal{F}[c(x,t) - \bar{c}(t)](q)|^{2}. \tag{S5}$$

Here, $\mathcal{F}$ is the two-dimensional discrete Fourier transform, $q = (q_x, q_y)$ is the Fourier wavevector and $q = |q|$ is its magnitude. The discrete wavenumbers were obtained from the FFT frequencies using the grid spacings $\Delta x$ and $\Delta y$.

The characteristic length was defined from the first moment of the structure factor, not from the peak wavenumber:

$$L_c(t) = \frac{2\pi}{\frac{\left[\sum_{q>0} q\ S(q,t)\right]}{\left[\sum_{q>0} S\,(q,t)\right]} + \epsilon}. \tag{S6}$$

The sums exclude the zero Fourier mode, and $\epsilon$ is a small numerical constant used to avoid division by zero. In time-series plots, the characteristic-length error was reported as the absolute difference between the predicted and reference $L_c(t)$. In the summary bars, it was reported as the mean relative error over the extrapolation interval.

To compare the full spectral morphology, the two-dimensional structure factor was also radially binned into 32 wavenumber shells and normalized by its total non-zero-mode power. The radial structure-factor error at each time point was the RMSE between the predicted and reference normalized radial distributions. For the summary bars, this RMSE was normalized by the root-mean-square magnitude of the reference radial distribution and then averaged over the extrapolation interval.

**Three-dimensional Allen-Cahn energetic consistency**

For the three-dimensional Allen-Cahn system, we evaluated whether the predicted rollouts preserved the energetic relaxation of the reference dynamics. The simulated system follows the 3D Allen-Cahn equation $u_t = \Delta u - (u^3 - u)$. The free energy was computed using the Ginzburg-Landau functional

$$E(t) = \sum_{i,j,k} \left[ \frac{1}{4} \left( u_{i,j,k}(t)^2 - 1 \right)^2 + \frac{1}{2} \left| \nabla_h u_{i,j,k}(t) \right|^2 \right] \Delta x \, \Delta y \, \Delta z. \quad \text{(S7)}$$

Here, $u_{i,j,k}(t)$ is the scalar order parameter at voxel $(i,j,k)$, $\Delta x$, $\Delta y$ and $\Delta z$ are the grid spacings, and $\nabla_h$ is the discrete spatial gradient. The gradient was evaluated by centered finite differences in all three directions with periodic boundary conditions. Both the bulk and interfacial contributions were computed as voxel-summed spatial integrals multiplied by the voxel volume $\Delta x \Delta y \Delta z$; no additional averaging over the number of voxels was applied.

The bulk contribution corresponds to the double-well potential $\frac{1}{4}(u^2 - 1)^2$, and the interfacial contribution corresponds to $\frac{1}{2}|\nabla_h u|^2$. For visualization, the free-energy trajectory was normalized by its initial value, $E(t)/E(0)$. For quantitative error calculation, the unnormalized free energy was used, and the reported free-energy error was the mean relative error with respect to the reference free energy over the extrapolation interval.

We also evaluated the ratio between the bulk and interfacial energy components:

$$R(t) = \frac{E_{\text{bulk}}(t)}{E_{\text{interface}}(t) + \epsilon}. \quad \text{(S8)}$$

This ratio measures whether the prediction preserves the balance between phase relaxation in the bulk and energy stored near interfaces. The bulk-to-interface ratio error was computed as the mean relative error between the predicted and reference ratios over the extrapolation interval.

**Hydrofoil wake spectral and vortical diagnostics**

For the hydrofoil-wake data, the predicted two-component velocity field was evaluated using spectral and vortical diagnostics. Let $u(x,t)$ and $v(x,t)$ denote the streamwise and transverse velocity components. Before computing the kinetic-energy spectrum, the spatial mean of each component was removed independently at each time step.

A two-dimensional FFT with orthonormal normalization was applied to the demeaned velocity components. For each Fourier mode $k$, the spectral kinetic-energy density used in the final metric table was

$$e(k,t) = \frac{1}{2} \frac{|\hat{u}'(k,t)|^2 + |\hat{v}'(k,t)|^2}{HW}. \quad \text{(S9)}$$

Here, $H$ and $W$ are the height and width of the velocity grid, $u'$ and $v'$ are the demeaned velocity components, and $\hat{u}'$ and $\hat{v}'$ are their Fourier coefficients. The radial kinetic-energy spectrum was obtained by summing $e(k,t)$ over radial wavenumber bins and averaging over the evaluated frames. In the final metric-table implementation, the radial-bin energy was not divided by the bin width.

The spectral error was defined as a relative $L_1$ discrepancy between the predicted and reference radial spectra:

$$\text{SE} = \frac{\sum_{m \in \mathcal{K}} | E_{\text{pred},m} - E_{\text{ref},m} |}{\sum_{m \in \mathcal{K}} E_{\text{ref},m} + \epsilon}. \quad \text{(S10)}$$

Here, $m$ indexes radial bins, $E_{\text{pred},m}$ and $E_{\text{ref},m}$ are the predicted and reference binned spectra, and $\mathcal{K}$ denotes the included bins. The first radial bin was excluded by default to reduce the influence of the lowest-wavenumber contribution. Thus, the final spectral error is a relative $L_1$ metric, not a log-spectrum RMSE or an $L_2$ error.

Finally, rotational wake structures were evaluated using the scalar vorticity

$$\omega(x,t) = \partial_x v(x,t) - \partial_y u(x,t). \quad \text{(S11)}$$

Spatial derivatives used centered finite differences in the domain interior and first-order one-sided differences at the boundaries. Vorticity correlation was the Pearson correlation coefficient between predicted and reference vorticity over all evaluated space-time samples, not a cosine similarity.

**Supplementary Note 4: Details of noise implementation**

To evaluate the robustness of our framework against noise contamination, Gaussian noise was added into the dataset. The noise addition process is defined by the following formula:

$$\mathrm{noise}_{\mathrm{data}} = \mathrm{truth} + R \cdot \sigma \cdot \frac{\max(\mathrm{truth}) - \min(\mathrm{truth})}{2} \qquad (S12)$$

where $R \sim \mathcal{N}(0,1)$ is a random variable sampled from a standard normal distribution, $\sigma$ is the noise intensity specified as a fraction of the data range (with values of $0.05$ for $5\%$, $0.1$ for $10\%$, and $0.2$ for $20\%$, see Supplementary Fig. 3), and $\max(\mathrm{truth})$ and $\min(\mathrm{truth})$ represent the maximum and minimum values of the ground truth data. This approach ensures that the added noise is proportional to the data range, effectively simulating realistic measurement noise while maintaining precise control over the noise intensity.

**Supplementary Note 5: Comparison with existing symbolic regression methods**

We compare our hierarchical physics-embedded framework with several representative symbolic regression approaches, including PeRCNN, sparse regression methods such as SINDy [7] and PDE-FIND [8], and end-to-end equation discovery techniques such as deep symbolic regression (DSR) [9] and genetic programming (GP) [10].

PeRCNN combines physics-encoded convolutional architectures with sparse regression to identify governing equations from spatiotemporal data. However, in complex systems with coupled nonlinear dynamics—such as the Cahn–Hilliard equation—PeRCNN often suffers from high reconstruction error. This is primarily because the physical terms must first be approximated through a full-field solution reconstruction, and any inaccuracies in this step inevitably propagate into the subsequent sparse regression, degrading the quality of discovered equations.

Furthermore, the sparse regression stage itself presents inherent challenges. It relies on predefined symbolic libraries and assumes that the target physical terms can be expressed as linear combinations of candidate functions. This not only requires substantial prior knowledge but also limits the model's capacity to recover non-polynomial or compositional expressions. Its performance deteriorates especially under sparse or noisy measurements, which are typical in real-world scenarios.

End-to-end approaches such as DSR and GP eliminate the reliance on explicit reconstruction steps, instead attempting to directly infer the governing equations from raw data. However, these methods lack the ability to disentangle multiple coupled physical effects. They typically require complete, high-resolution datasets to achieve stable training and meaningful results. Without effective integration of known physical knowledge, their search space remains large, and optimization often results in unstable, inaccurate, or overly complex expressions.

Our framework addresses these limitations through a hierarchical and decoupled discovery process. By first reconstructing spatiotemporal dynamics using physics-embedded learning, we can extract intermediate physical terms that are structurally aligned with the PDE form. These terms are then individually mapped to symbolic expressions using deep symbolic regression guided by reinforcement learning. This two-stage design avoids the error amplification seen in PeRCNN, eliminates the reliance on symbolic libraries, and decouples the discovery of multiple physical terms. Importantly, our method remains effective under sparse and noisy conditions, achieving robust recovery of governing equations even in the absence of dense measurements.

**Supplementary Note 6: Proof of Theorem 1, Theorem 2 and Theorem 3**

**Proof of Theorem 1.** By Assumption 1, for any $\boldsymbol{u} \in \mathcal{U}$, we have:

$$\left\|\hat{\mathcal{F}}(\boldsymbol{u},\theta) - \mathcal{F}(\boldsymbol{u})\right\|_2 = \left\|\hat{\mathcal{C}}\left(\hat{\mathcal{E}}(\boldsymbol{u},\theta_{\hat{\mathcal{E}}}),\theta_{\hat{\mathcal{C}}}\right) - \left(\mathcal{C}^*\left(\mathcal{E}^*(\boldsymbol{u})\right) + \mathcal{R}(\boldsymbol{u})\right)\right\|_2$$

$$\leq \left\|\mathcal{C}^*\left(\hat{\mathcal{E}}(\boldsymbol{u},\theta_{\hat{\mathcal{E}}})\right) - \mathcal{C}^*\left(\mathcal{E}^*(\boldsymbol{u})\right)\right\|_2 + \left\|\hat{\mathcal{C}}\left(\hat{\mathcal{E}}(\boldsymbol{u},\theta_{\hat{\mathcal{E}}}),\theta_{\hat{\mathcal{C}}}\right) - \mathcal{C}^*\left(\hat{\mathcal{E}}(\boldsymbol{u},\theta_{\hat{\mathcal{E}}})\right)\right\|_2 + \|\mathcal{R}(\boldsymbol{u})\|_2. \quad \text{(S13)}$$

For the first term in equation (S13), Assumption 2 implies:

$$\left\|\mathcal{C}^*\left(\hat{\mathcal{E}}(\boldsymbol{u},\theta_{\hat{\mathcal{E}}})\right) - \mathcal{C}^*\left(\mathcal{E}^*(\boldsymbol{u})\right)\right\|_2 \leq L_{\mathcal{C}}\left\|\hat{\mathcal{E}}(\boldsymbol{u},\theta_{\hat{\mathcal{E}}}) - \mathcal{E}^*(\boldsymbol{u})\right\|_2$$

$$= L_{\mathcal{C}}\left(\sum_{i=1}^{m}\left\|\hat{\mathcal{E}}_i(\boldsymbol{u},\theta_{\hat{\mathcal{E}}}) - \mathcal{E}_i^*(\boldsymbol{u})\right\|_2^2\right)^{\frac{1}{2}}. \quad \text{(S14)}$$

For the definition of $\epsilon_{\mathcal{E}_i}$, $\epsilon_{\mathcal{C}}$ and $\delta_{\text{comp}}$, we have $L_{\mathcal{C}}\left(\sum_{i=1}^{m}\left\|\hat{\mathcal{E}}_i(\boldsymbol{u},\theta_{\hat{\mathcal{E}}}) - \mathcal{E}_i(\boldsymbol{u})\right\|_2^2\right)^{\frac{1}{2}} \leq L_{\hat{\mathcal{C}}}\left(\sum_{i=1}^{m}\epsilon_{\mathcal{E}_i}^2\right)^{\frac{1}{2}}$, $\left\|\hat{\mathcal{C}}\left(\hat{\mathcal{E}}(\boldsymbol{u},\theta_{\hat{\mathcal{E}}}),\theta_{\hat{\mathcal{C}}}\right) - \mathcal{C}^*\left(\hat{\mathcal{E}}(\boldsymbol{u},\theta_{\hat{\mathcal{E}}})\right)\right\|_2 \leq \epsilon_{\mathcal{C}}$ and $\|\mathcal{R}(\boldsymbol{u})\|_2 \leq \delta_{\text{comp}}$, respectively. Therefore, we have:

$$\left\|\hat{\mathcal{F}}(\boldsymbol{u},\theta) - \mathcal{F}(\boldsymbol{u})\right\|_2 \leq \epsilon_{\mathcal{C}} + L_{\mathcal{C}}\left(\sum_{i=1}^{m}\epsilon_{\mathcal{E}_i}^2\right)^{\frac{1}{2}} + \delta_{\text{comp}}. \quad \text{(S15)}$$

Taking the supremum of $\left\|\hat{\mathcal{F}}(\boldsymbol{u},\theta) - \mathcal{F}(\boldsymbol{u})\right\|_2$ over $\boldsymbol{u} \in \mathcal{U}$ yields Theorem 1. ■

**Proof of Theorem 2.** By definitions of $\mathcal{L}(u)$ and $\mathcal{L}_K^*(u)$, we have:

$$\mathcal{L}(u)(\boldsymbol{x}) - \mathcal{L}_K^*(u)(\boldsymbol{x}) = \sum_{\|\boldsymbol{\xi}\|_\infty > K} \sigma_{\mathcal{L}}(\boldsymbol{\xi})\hat{u}(\boldsymbol{\xi})e_{\boldsymbol{\xi}}(\boldsymbol{x}). \quad \text{(S16)}$$

Using Parseval's identity under the Fourier convention adopted in Section 5.3, we obtain

$$\|\mathcal{L}(u) - \mathcal{L}_K^*(u)\|^2 = \sum_{\|\boldsymbol{\xi}\|_\infty > K} |\sigma_{\mathcal{L}}(\boldsymbol{\xi})|^2|\hat{u}(\boldsymbol{\xi})|^2. \quad \text{(S17)}$$

By Assumption 3, we have

$$\|\mathcal{L}(u) - \mathcal{L}_K^*(u)\|^2 \leq C_{\mathcal{L}}^2 \sum_{\|\boldsymbol{\xi}\|_\infty > K} \left(1 + \|\boldsymbol{\xi}\|_\infty\right)^{2p} |\hat{u}(\boldsymbol{\xi})|^2. \quad \text{(S18)}$$

For every frequency satisfying $\|\boldsymbol{\xi}\|_\infty > K$, we have

$$\left(1 + \|\boldsymbol{\xi}\|_\infty\right)^{2p} = \left(1 + \|\boldsymbol{\xi}\|_\infty\right)^{-2r}\left(1 + \|\boldsymbol{\xi}\|_\infty\right)^{2(p+r)} \leq (1+K)^{-2r}\left(1 + \|\boldsymbol{\xi}\|_\infty\right)^{2(p+r)}. \quad \text{(S19)}$$

Substituting this bound into the inequality (S18) gives

$$\|\mathcal{L}(u) - \mathcal{L}_K^*(u)\|^2 \leq C_{\mathcal{L}}^2(1+K)^{-2r} \sum_{\|\boldsymbol{\xi}\|_\infty > K} \left(1 + \|\boldsymbol{\xi}\|_\infty\right)^{2(p+r)} |\hat{u}(\boldsymbol{\xi})|^2. \quad \text{(S20)}$$

Since the truncated tail is bounded by the full spectral sum, Assumption 3 implies

$$\sum_{\|\boldsymbol{\xi}\|_\infty > K} \left(1 + \|\boldsymbol{\xi}\|_\infty\right)^{2(p+r)} |\hat{u}(\boldsymbol{\xi})|^2 \leq \sum_{\boldsymbol{\xi} \in \mathbb{Z}^d} \left(1 + \|\boldsymbol{\xi}\|_\infty\right)^{2(p+r)} |\hat{u}(\boldsymbol{\xi})|^2 \leq M_{\mathcal{U}}^2. \quad \text{(S21)}$$

Thus

$$\|\mathcal{L}(u) - \mathcal{L}_K^*(u)\|^2 \leq C_{\mathcal{L}}^2 M_{\mathcal{U}}^2 (1+K)^{-2r}. \quad \text{(S22)}$$

Taking the square root yields

$$\|\mathcal{L}(u) - \mathcal{L}_K^*(u)\| \leq C_{\mathcal{L}} M_{\mathcal{U}} (1+K)^{-r}. \quad \text{(S23)}$$

Because the above estimate holds for every $u \in \mathcal{U}$, we obtain the uniform truncation bound

$$\sup_{u \in \mathcal{U}} \|\mathcal{L}(u) - \mathcal{L}_K^*(u)\| \leq C_{\mathcal{L}} M_{\mathcal{U}} (1+K)^{-r}. \quad \text{(S24)}$$

This proves the approximation-error estimate in Theorem 2. It remains to derive the parameter-scaling law. To guarantee a prescribed component accuracy $\eta > 0$, it is sufficient to require

$$C_{\mathcal{L}} M_{\mathcal{U}} (1+K)^{-r} \leq \eta. \quad \text{(S25)}$$

Equivalently,

$$K \geq \left(\frac{C_{\mathcal{L}} M_{\mathcal{U}}}{\eta}\right)^{\frac{1}{r}} - 1. \quad \text{(S26)}$$

Thus, it is sufficient to take

$$K = \left\lceil \left( \frac{C_{\mathcal{L}} M_{\mathcal{U}}}{\eta} \right)^{\frac{1}{r}} - 1 \right\rceil. \tag{S27}$$

to ensure $\|\mathcal{L}(u) - \mathcal{L}_K^*(u)\| \leq \eta$. ■

**Proof of Theorem 3.** We prove the parameter-complexity bound by combining the hierarchical error decomposition with the component-wise Fourier parameter-scaling result in Theorem 2.

From the hierarchical error decomposition in Theorem 1, the approximation error of HPE-AFNO satisfies

$$\epsilon_{\text{total}} \leq \epsilon_{\mathcal{C}} + L_{\mathcal{C}} \left( \sum_{i=1}^{m} \epsilon_{\mathcal{E}_i}^2 \right)^{\frac{1}{2}} + \delta_{\text{comp}}. \tag{S28}$$

where $\epsilon_{\mathcal{E}_i}$ is the approximation error of the $i$-th first-level physical-expression extractor, $\epsilon_{\mathcal{C}}$ is the approximation error of the second-level governing combiner, $L_{\mathcal{C}}$ is the Lipschitz constant of the ideal combiner, and $\delta_{\text{comp}}$ is the compositional residual.

Let the prescribed global accuracy be $\epsilon > 2\delta_{\text{comp}}$. We allocate the approximation budget as

$$\epsilon_{\mathcal{C}} \leq \frac{\epsilon}{4}, \tag{S29}$$

and for each $i = 1,2,\cdots,m$

$$\epsilon_{\mathcal{E}_i} \leq \frac{\epsilon}{4 L_{\mathcal{C}} \sqrt{m}}. \tag{S30}$$

Moreover, because $\epsilon > 2\delta_{\text{comp}}$, we have

$$\delta_{\text{comp}} \leq \frac{\epsilon}{2}. \tag{S31}$$

Therefore

$$\epsilon_{\text{total}} \leq \frac{\epsilon}{4} + L_{\mathcal{C}} \left( \sum_{i=1}^{m} \left( \frac{\epsilon}{4 L_{\mathcal{C}} \sqrt{m}} \right)^2 \right)^{\frac{1}{2}} + \frac{\epsilon}{2} = \epsilon. \tag{S32}$$

Thus, it is sufficient to approximate the second-level combiner to accuracy $\frac{\epsilon}{4}$ and each first-level extractor to accuracy $\frac{\epsilon}{4 L_{\mathcal{C}} \sqrt{m}}$.

According to Assumption 4, approximating the $i$-th extractor to accuracy $\eta_{\mathcal{E}_i} = \frac{\epsilon}{4 L_{\mathcal{C}} \sqrt{m}}$ requires a sufficient number of learnable Fourier parameters bounded by

$$N_{\mathcal{E}_i}(\eta_{\mathcal{E}_i}) \leq \alpha_{\mathcal{E}_i} \eta_{\mathcal{E}_i}^{-\frac{d}{r_{\mathcal{E}_i}}} = \alpha_{\mathcal{E}_i} \left( \frac{4 L_{\mathcal{C}} \sqrt{m}}{\epsilon} \right)^{\frac{d}{r_{\mathcal{E}_i}}}. \tag{S33}$$

Similarly, approximating the second-level combiner to accuracy $\eta_{\mathcal{C}} = \frac{\epsilon}{4}$, requires at most

$$N_{\mathcal{C}}(\eta_{\mathcal{C}}) \leq \alpha_{\mathcal{C}} \eta_{\mathcal{C}}^{-\frac{d}{r_{\mathcal{C}}}} = \alpha_{\mathcal{C}} \left( \frac{4}{\epsilon} \right)^{\frac{d}{r_{\mathcal{C}}}}. \tag{S34}$$

learnable Fourier parameters. The total number of learnable Fourier parameters in the hierarchical model is the sum of the parameters used by the first-level extractors and the second-level combiner. Therefore,

$$N_{\text{HPE}}(\epsilon) \leq \sum_{i=1}^{m} \alpha_{\mathcal{E}_i} \left( \frac{4 L_{\mathcal{C}} \sqrt{m}}{\epsilon} \right)^{\frac{d}{r_{\mathcal{E}_i}}} + \alpha_{\mathcal{C}} \left( \frac{4}{\epsilon} \right)^{\frac{d}{r_{\mathcal{C}}}}. \tag{S35}$$

This proves the hierarchical parameter-complexity bound. ■